\documentclass[sn-mathphys,Numbered]{sn-jnl}

\usepackage{graphicx}%
\usepackage{amsmath}%
\usepackage{amssymb}%
\usepackage{mathrsfs}%
\usepackage[mathscr]{euscript}
\usepackage[title]{appendix}%
\usepackage{xcolor}%
\usepackage{textcomp}%
\usepackage{manyfoot}%
\usepackage{booktabs}%
\usepackage{algorithm}%
\usepackage{algorithmicx}%
\usepackage{algpseudocode}%
\usepackage{listings}%

\usepackage[utf8]{inputenc}
\usepackage{epstopdf}
\usepackage[normalem]{ulem}
\usepackage{bm}
\usepackage{url}
\usepackage{esvect}
\usepackage{cases}
\usepackage{subfigure}
\usepackage{multirow}

\usepackage{lmodern}

\usepackage{placeins}

\usepackage{xcolor}
\newcommand{\REV}[1]{{\color{black}#1}}

\newtheorem{problem}{Problem}

\begin{document}

\title[Article Title]{A Schwarz-Christoffel Mapping-based Framework for Sim-to-Real Transfer in Autonomous Robot Operations}


\author*[1]{\fnm{Shijie} \sur{Gao}}\email{sjgao@virginia.edu}

\author[1]{\fnm{Nicola} \sur{Bezzo}}\email{nbezzo@virginia.edu}

\affil[1]{\orgdiv{Charles L. Brown Department of Electrical and Computer Engineering}, \orgname{University of Virginia}, \orgaddress{\street{}, \city{Charlottesville}, \postcode{22903}, \state{VA}, \country{USA}}}


\abstract{
Despite the remarkable acceleration of robotic development through advanced simulation technology, robotic applications are often subject to performance reductions in real-world deployment due to the inherent discrepancy between simulation and reality, often referred to as the ``sim-to-real gap". This gap arises from factors like model inaccuracies, environmental variations, and unexpected disturbances. Similarly, model discrepancies caused by system degradation over time or minor changes in the system's configuration also hinder the effectiveness of the developed methodologies. Effectively closing these gaps is critical and remains an open challenge. This work proposes a lightweight conformal mapping framework to transfer control and planning policies from an expert teacher to a degraded less capable learner. The method leverages Schwarz-Christoffel Mapping (SCM) to geometrically map teacher control inputs into the learner's command space, ensuring maneuver consistency. To demonstrate its generality, the framework is applied to two representative types of control and planning methods in a path-tracking task: 1) a discretized motion primitives command transfer and 2) a continuous Model Predictive Control (MPC)-based command transfer. The proposed framework is validated through extensive simulations and real-world experiments, demonstrating its effectiveness in reducing the sim-to-real gap by closely transferring teacher commands to the learner robot.

}





\maketitle
\section{Introduction}\label{sec:introduction}
    In recent years, the advancements in simulation technologies have led to a significant surge in robotic research and applications~\cite{yan2020close,collins2021review}. Simulations provide a low-cost solution, virtual proving ground for designing and controlling robots, allowing for rapid prototyping and testing without associated risks~\cite{afzal2020study}. However, despite the flawless performance of behaviors and algorithms in simulated settings, the ``reality gap" between simulated and real environments and inherent discrepancies in robotic models often lead to performance degradation or even failures when directly applying these well-developed techniques
    in the real world. Despite considerable time and resources devoted to creating techniques within simulations, researchers still face formidable challenges when applying these methods to specific platforms in the real world. Thus, closing this gap is essential for advancing the practical deployment of robotic systems in diverse fields such as soft robotics applications~\cite{lipson2014challenges} and agriculture~\cite{rizzardo2020importance}.

\begin{figure}[ht]
        \centering
        \includegraphics[width=0.7\columnwidth]{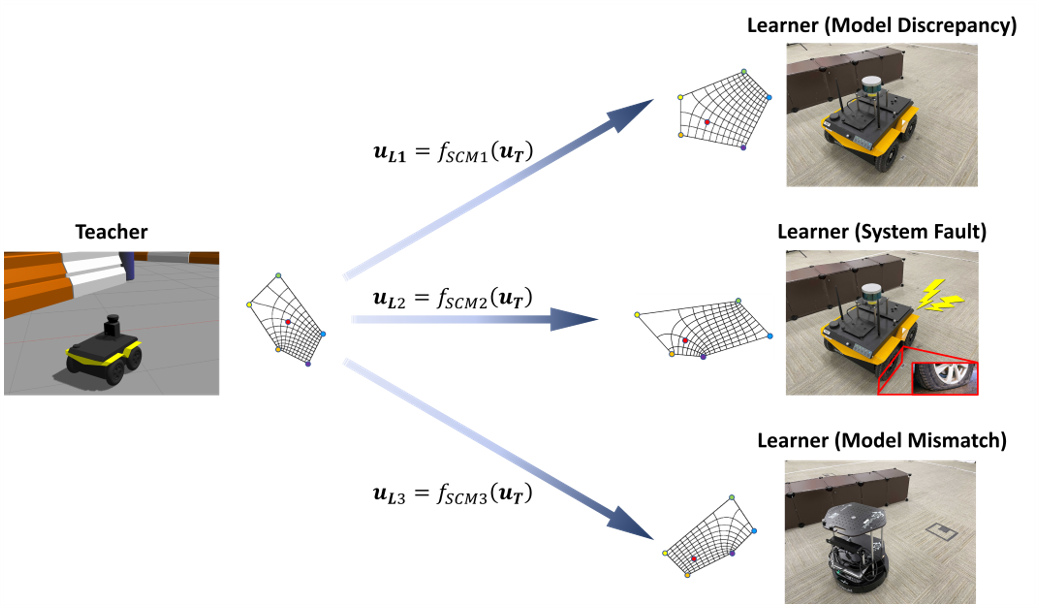}
        \caption{A pictorial representation of the proposed transfer method. A desired command is transferred from an expert teacher to a learner, which may differ from the teacher due to unmodeled dynamics, failures, disturbances and platform aging.}
        \label{fig:Intro}
    \end{figure}

    Moreover, understanding how to effectively bridge this gap contributes to solving broader domain-transfer problems. For instance, consider the impact of system aging. Although systems before and after aging are different stages of the same system with fundamentally similar dynamics, a velocity command that once effectively propelled the vehicle might result in diminished speed due to wear and tear. This type of gap, while distinct from the sim-to-real discrepancy, belongs to the broader category of model mismatches discrepancy. Similar problems are not limited only to mechanical aging, but can also be found when dealing with environment changes, external disturbances, software bugs, and even failures that deprecate and modify the system's original model. In this paper, we seek a general framework to transfer and adapt the system's performance from one vehicle to another. The goal of the proposed work is to: 
    \begin{itemize}
        \item reduce the sim-to-real gap allowing a developer to quickly transfer motion planning and control methods onto a real platform;
        \item transfer knowledge designed for a specific robot to a similar robot;
        \item compensate for system deterioration/failures by quickly learning the limits and the proper input mapping to continue an operation.
    \end{itemize}
    All of the aforementioned problems can be simplified as a {\em teacher} transferring the control and motion planning policies to a {\em learner}. To address this transferring problem we introduce a lightweight, conformal mapping-based transfer learning framework as depicted in Figure~\ref{fig:Intro}. The proposed framework maps directly the control inputs of teachers to learner systems, avoiding learning their dynamic models. The framework also learns and considers the learner's limits, so that the transferred motion plan is achievable by mapped control inputs. 

    In this work, we aim to provide a robust solution to the pervasive problem of model mismatch in robotics, enhancing both the efficacy and efficiency of deploying robotic applications. Overall, the contribution of this work is threefold: 

    \newpage
    \begin{enumerate}
    \item \textbf{Control Inputs Transferring}: We present a lightweight transfer framework, utilizing Schwarz-Christoffel Mapping (SCM) theory for the direct transfer of control inputs from the teacher to the learner. This innovative approach allows the learner to adopt the teacher's control policies without the need to understand its own dynamics.
    \end{enumerate}

    Within our proposed transfer framework, we demonstrate such transferring with two distinct types of control and motion planning methods: 
    
    \begin{enumerate}
    \item[2.] \textbf{Motion Primitive-based Transfer}: We leverage the proposed transfer framework to first identify the teacher's motion primitives that the learner can execute, and then transfer the associated discrete commands to the learner, enabling replication of the teacher's motions.
    
    \item[3.] \textbf{Model Predictive Control Transfer}: We incorporate a Model Predictive Controller (MPC) as a unified teacher controller and path planner within the proposed transfer framework. The framework imposes constraints on the MPC to ensure that the optimized control input remains within the learner's operational limits. These inputs are then mapped to the learner, allowing it to mirror the teacher's movements in a continuous control space.
    
    \end{enumerate}
    In addition to the contributions mentioned above, we validate the proposed sim-to-real mitigation framework with extensive simulations and real-world experiments. To the best of our knowledge, this work is novel in leveraging the conformal mapping method for transferring control and motion planning policies between robotic systems. Distinct from existing literature, our work bridges the sim-to-real gap by focusing exclusively on the domain of control inputs, rather than attempting to learn specific model discrepancies. By directly translating control inputs, we can achieve a lighter-weight seamless transfer that avoids the need for extensive data collection and training to learn the precise model mismatch -- which may not be always possible during a mission. 

\section{Related Work}\label{sec:related_work}
In this section, we review the state-of-the-art transfer learning methods in the current literature, especially those that focus on domain migration in robotic applications and sim-to-real problems. Subsequently, we discuss the conformal mapping techniques used in robotic applications.

\subsection{Sim-to-Real Transferring}
     The sim-to-real gap of transferring from the simulation to the real world exists mainly because either the model is not accurate or the environmental factors do not appear in the simulation. \REV{One of the main approaches to mitigate the sim-to-real gap or adapt to dynamic changes is to improve simulation/model fidelity, ensuring the agent is trained in an environment that is as realistic as possible. Several methods leverage this idea. For example, \cite{jeong2019modelling} performs system identification to align new or simulated dynamics with reality and adds learnable models for additional inputs, \cite{kaspar2020sim2real} builds realistic training procedures that mimic how humans learn in the physical world, \cite{hundt2020good} uses reward shaping techniques to encourage greater exploration of correct behaviors, while \cite{nachum2019multi} propose a hierarchical learning method by providing atomic actions that can be combined to create complex tasks. However, these methods are computationally expensive, especially if used online \cite{truong2023rethinking}. They are also not generalizable as their abstractions target specific robots in specific environment configurations. In contrast, our proposed framework does not require extracting precise dynamics from the actual system. Instead, it leverages a simple abstraction model, enabling broader generalization and faster transfer.} Although we can always improve the fidelity of the simulation according to the real-world observations~\cite{chebotar2019closing}, it is yet not applicable to perfectly replicate the reality and may lose generality when deploying applications developed in such simulators as pointed out in~\cite{truong2023rethinking}. Thus, apart from polishing the simulator, it is necessary to develop a more robust controller and planner or other transferring method to close the sim-to-real gap. 
     
     As machine learning techniques have become widely exploited, transfer learning has gained significant attention in robotics toward addressing transferring problem. The core concept of transfer learning is to migrate the knowledge gained from one task to address similar tasks in diverse environments. It takes advantage of existing knowledge~\cite{devin2017learning} and reduces the cost as well as the risk associated with data collection and model training~\cite{fremont2020formal,zhang2020cautious, james2019sim}. Researchers have been investigating this topic through various strategies, including domain randomization, domain adaptation, imitation learning, and large fundamental models etc. 
     
     Domain randomization approaches parameterize the dynamics and environmental conditions within simulations, seeking to encapsulate real-world complexities through augmented simulation scenarios~\cite{peng2018sim,tobin2017domain, tobin2018domain,chebotar2019closing}. Despite the intention to bridge the sim-to-real gap, they often incur high computational costs and can lead to over-generalization, resulting in systems that are capable of handling a broad range of unlikely scenarios but may underperform in typical real-world conditions. \REV{Additionally, in contrast to our approach, this type of approach relies on strong assumptions about the system model structure, which requires a similar parametrization between simulated and real robots. Our approach overcomes these limitations by directly transferring control inputs and introducing perturbations only when needed.} Domain adaptation, a well-established technique in machine learning, trains models with samples from a source domain to effectively generalize to a target domain. \REV{It is applicable in situations where there is little prior knowledge about the environment, and need to explore it by iteratively polishing the learned agent based on incremental knowledge \cite{van2000statistics}. Meta learning \cite{yel2022meta} is another technique that has been widely used for domain adaptation where a model is learned in the real world or simulation, and a secondary model is generated from the primary model to adapt to new behaviors, based on observed data in the physical world \cite{arndt2020meta}. Our proposed method has some similarities in the problem structure; however, we do not rely on fine-tuned learning components and extensive training data for transfer.} Proven in computer vision, Transfer Learning method has been particularly useful for bridging the sim-to-real gap in vision-based robotic control problems\cite{bousmalis2018using, fang2018multi}. Similarly, recent end-to-end approaches\cite{truong2023i2o} and RL-CycleGAN \cite{rao2020rl} also focus on minimizing the sim-to-real disparity primarily in visual-based control tasks by aligning the visual inputs between reality and simulations. However, these approaches often overlook the underlying model mismatches and are built on the assumption that robotic controls are well-established without sim-to-real discrepancies. This assumption represents a significant oversight—a challenge that our work aims to address.
     
     Another strategy that has been broadly explored is imitation learning, also known as learning from demonstration or behavior cloning, which enables robots to acquire new skills by mimicking expert behaviors. This method shifts from the traditional approach of learning through prolonged repetition of simple tasks by directly deriving policies~\cite{shavit2018learning,kormushev2011imitation}, plans~\cite{kulic2012incremental,niekum2012learning}, or rewards~\cite{bajcsy2017learning, finn2016guided} from an expert. While this technique offers some performance guarantees, it can be adversely affected by suboptimal or inappropriate examples from the expert, and it lacks the ability to handle complex tasks. Additionally, it is sensitive to environmental changes, which can hinder its ability to generalize effectively~\cite{ravichandar2020recent}. Consequently, imitation learning is not ideally suited for addressing sim2real transfer problems, where adaptability across varied real-world conditions is crucial. Recently, large foundation models have been adapted for robotics applications, exemplified by~\cite{brohan2022rt, brohan2023rt}. These works have shown their effectiveness in high-level semantic reasoning using visual and language inputs. While they enhance human-robot interaction and enable the transfer of high-level plans between different robots and environments, their application is limited to tasks for which the robots have been explicitly trained. Although the potential for transferring control through transformers remains largely unexplored, the extensive data requirements for training such models pose a significant challenge for researchers.

    
    \subsection{Conformal Mapping}
    
    Conformal mapping is a mathematical technique used in complex analysis. A conformal mapping function transfers a complex domain onto another one while preserving angles locally. Researchers have exploited this approach for geometrical problems in robotics. \cite{bayro2006conformal} has introduced conformal mapping to aid in correcting the distortion in robotic vision. \cite{kosari2017using, song2019distortion} treat the original path plan as a geometrical pattern and use conformal mapping to adapt the plan to suit the task in specific settings. However, the SCM method proposed in this paper is rarely utilized in the robotics field. In~\cite{notomista2018coverage}, the SCM is employed to transform planar motion into continuous linear motion, addressing a coverage control problem for wire-traversing robots. Our previous work~\cite{gao2021conformal} first brought the idea of leveraging the SCM method to directly transfer control inputs between two systems. It shows the efficiency and effectiveness of bridging the gap when transferring the path planner and the controller between similar systems. However, the previous work relies on motion primitives to achieve motion plan transferring, whose transferring results are discrete and highly depend on the size of the motion primitive library. In this work, we enhance the generalizability and robustness of the previously proposed transfer framework by 1) eliminating the need for a dedicated calibration stage, thereby enabling simultaneous learning of the learner's limits and transferring of control and path plan; 2) fully leveraging the learner's capabilities to allow transfers over a continuous command space; 3) strengthening the framework's ability to handle system process noises and environmental uncertainties.

\section{Problem Formulation}\label{sec:problem_formulation}
The problem addressed in this work can be considered as a transfer learning problem from the teacher system to the learner. The goal is to find a mapping function that allows for a seamless translation of commands from the teacher to the learner, ensuring they result in identical movements (i.e., reaching the same pose and speed). The user possesses comprehensive knowledge of the teacher including its dynamics, finely-tuned controller, and planner. In contrast, the learner is treated as a black box to which we can send control inputs and observe the full states. 

\vspace{5pt}
\noindent
\textbf{Notations - }In this work, we use $\bm{x_T}(t)$ and $\bm{x_L}(t)$ to represent the state of the teacher and learner systems while $\bm{u_T}(t)\subset\mathbb{R}^2$ and $\bm{u_L}(t)\subset\mathbb{R}^2$ represent the control inputs of the two systems at time $t$. The symbols $\overline{\bm{u}}$ and $\underline{\bm{u}}$ denote the upper and lower bounds of the control inputs, respectively. 

Formally, we define two problems in this context:

\vspace{4pt}
\begin{problem}{\bf{\emph{Teacher-Learner Control Transfer:}}}\label{problem1}
Given a teacher robot with dynamics ${\bm{x_T}(t+1)}{=}f_T(\bm{x_T}(t),\bm{u_T}(t))$ and control law $\bm{u_{T}}{=}g(\bm{x})$, find a policy to map $\bm{u_{T}}$ to a learner input $\bm{u_{L}}$ such that $\bm{x_{L}}(t+1){=}f_L(\bm{x_L}(t),\bm{u_L}(t))\approx f_T(\bm{x_L}(t),\bm{u_T}(t))$, without knowing the closed-form of $f_L$.
\end{problem}

\vspace{4pt}
\begin{problem}{\bf{\emph{Teacher-Learner Motion Planning Adaptation:}}}\label{problem2}
Consider a task to navigate from an initial location to a final goal $\bm{x}_G$. Assume that the learner's input space $\bm{u_L}\in[\bm{\underline{u}_L}, \bm{\overline{u}_L}] \subset [\bm{\underline{u}_T}, \bm{\overline{u}_T}]$; design a motion planning policy $\bm{\pi_{T}}^{L}$ for the teacher that considers the limitations of the learner and such that the desired trajectory calculated, $\tau$, can be tracked by the learner, i.e., such that $\|\bm{x}_L-\bm{x}_{\tau}\|\leq \epsilon$ where $\epsilon$ is a small deviation threshold.
\end{problem}

We assume that the learner has similar kinematics as the teacher but is less capable. By ``similar", we mean that they share the same configuration space $\mathscr{C}_L = \mathscr{C}_T$ and task space $\mathscr{T}_L = \mathscr{T}_T$, and their kinematic models exhibit a fundamental similarity in structure and governing equations. Although differences in parameters or external influences underline their distinctions, a shared mathematical foundation emphasizes their conceptual similarity. This similarity provides a basis for applying methodologies across both systems. For example, both systems might be based on the same dynamic models but with distinct parameter sets, or they share identical parameters that fail to precisely reflect the actual behavior of the learner system.  Regarding ``less capability", this implies that the learner's command space is a subset of the teacher's $\bm{U}_L \subset \bm{U}_T$. For instance, the learner may not be able to drive as fast or turn as sharply as the teacher. This allows the teacher to perform all the learner's maneuvers but not vice versa. The assumption aligns with our focus on transferring knowledge from the simulated system into a real-world vehicle, as a virtual system can often be designed to surpass the limits of its real counterpart in sim-to-real problems. 

\section{Methodology}\label{sec:framework}
\begin{figure}[ht]
        \centering
        \includegraphics[width=0.8\columnwidth]{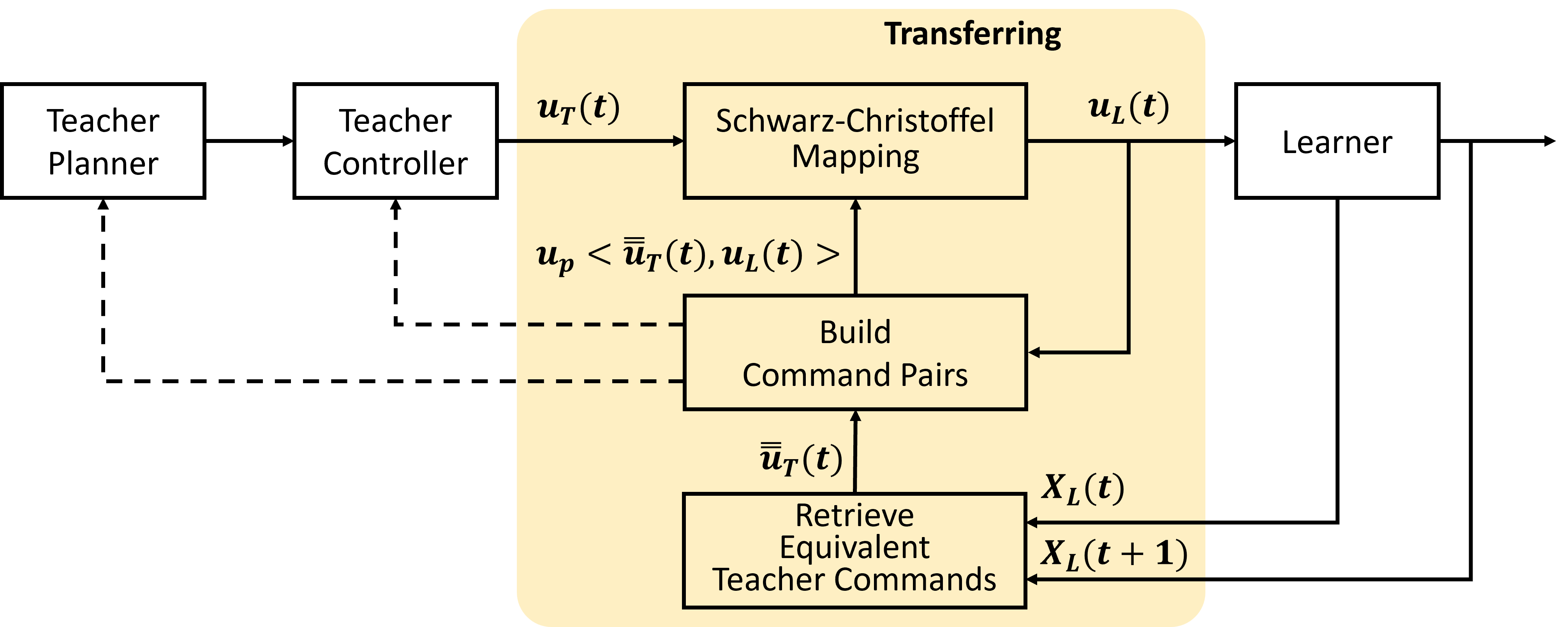}
        \caption{The block diagram of the proposed mapping-based transfer learning framework}
        \label{fig:overallFramework}
    \end{figure}

    To address the formalized problems, we propose a conformal-mapping-based transfer learning framework. The block diagram in Figure~\ref{fig:overallFramework} presents the architectural overview of the proposed framework. The highlighted blocks are the main components for transferring the control inputs. Problem~\ref{problem1} is solved by leveraging SCM to conformally map the teacher vehicle's control input to obtain the learner's command. To tackle Problem~\ref{problem2}, we learn how to constrain the teacher's command domain so that the teacher's control inputs, which are yielded from the teacher's control and planning policies, are tailored to accommodate the learner's limit after mapping. This section presents the details of the proposed framework, explaining how the two problems are solved.
    
    \subsection{Control Transfer}\label{subsubsec:SCM_control_transfer}

    \noindent
    \textbf{\emph{Command Pairing}}
    
    In order to derive the mapping function, a fundamental step is understanding the correspondence between a learner's command and its equivalent from the teacher. The teacher's equivalents can be derived by computing the inverse kinematics of the teacher over the observed learner's motions. By linking the learner's command $\bm{u_L}$ with the teacher's equivalent $\bm{\overline{\overline{u}}_T}$ that is capable of mirroring the learner's altered behavior, we can preserve the geometric information of the command domains to grasp the difference between the systems.  Formally, we call such interconnected commands a command pair which is defined as follows,
    
    \begin{equation}\label{eq:commandPairs}
        \begin{aligned}
            & \bm{u_{p}} = \langle\bm{\overline{\overline{u}}_T}, \bm{u_L}\rangle \\
            \text{s.t.} \quad \quad & f_T(\bm{x_L}(t),\bm{\overline{\overline{u}}_T}(t)) = f_L(\bm{x_L}(t),\bm{u_L}(t)) \\
            \text{where} \quad \quad & \bm{\overline{\overline{u}}_T} = f_{T}^{-1}(\bm{x_L}(t),\bm{x_L}(t+1))
        \end{aligned}
    \end{equation}
    
\begin{figure}[h]
        \centering
            \subfigure[]{
                \includegraphics[width = 0.45\textwidth]{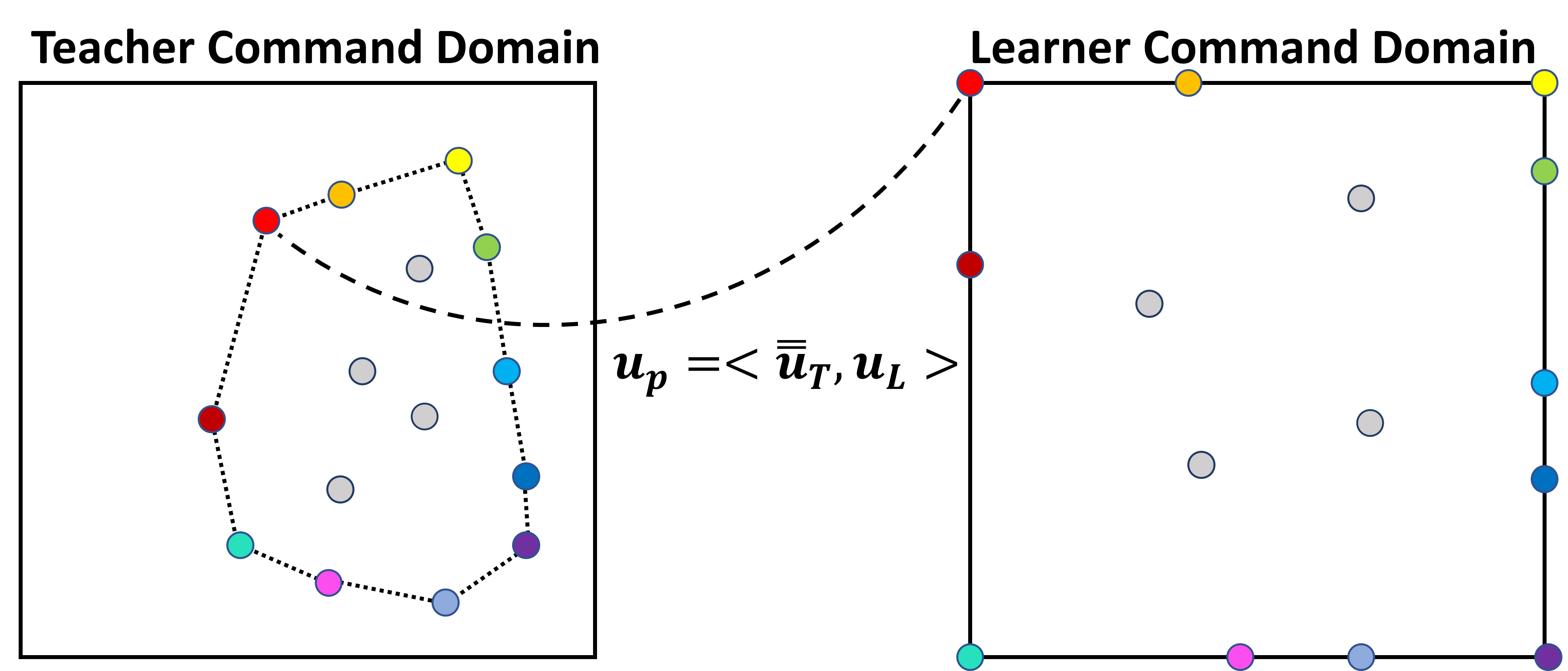}
                \label{fig:capability_discrete}
                }
            \hspace{5pt}
            \subfigure[]{
                \includegraphics[width =0.45\textwidth]{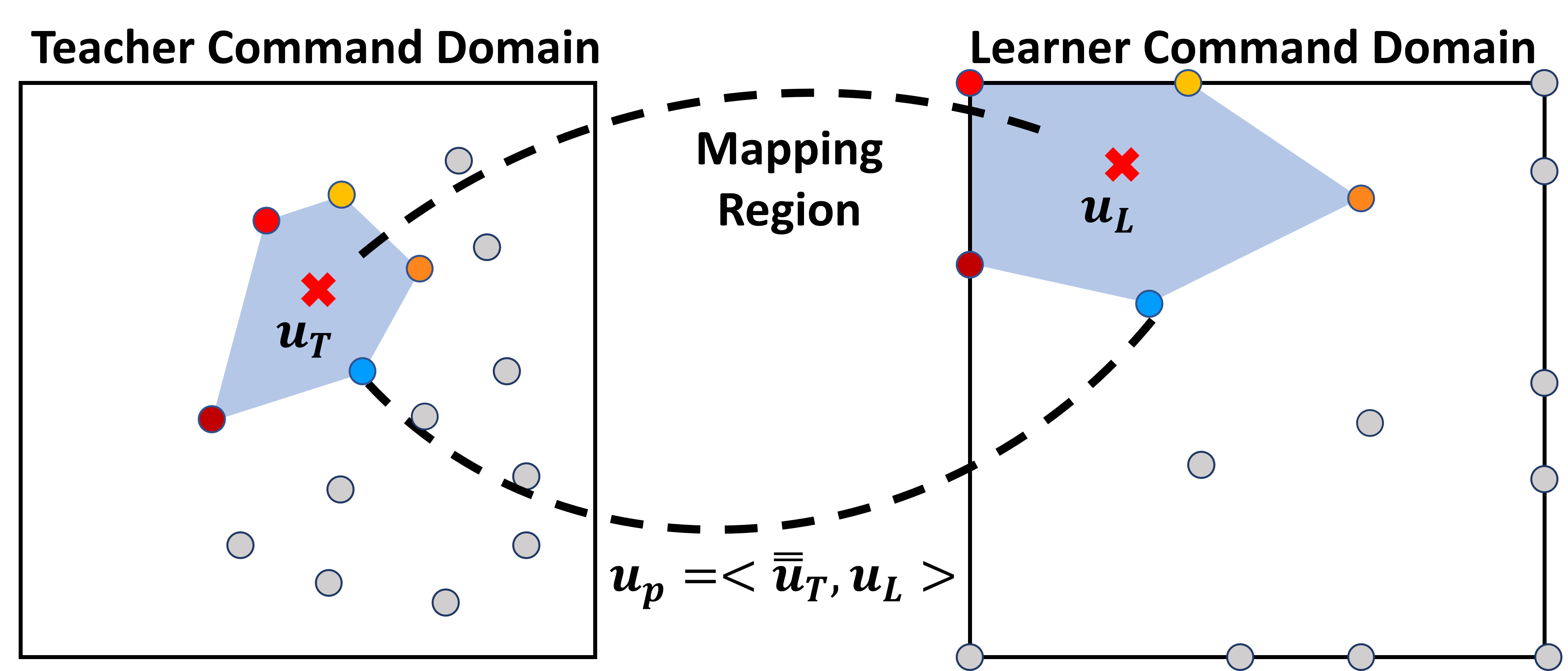}
                \label{fig:commandPairs_map}
            }
        \caption{(a) Command pairs are color-coded. The dashed envelope indicates the learner's limits on the teacher's command domain; (b) Selected command pairs for constructing the mapping regions are color-coded while the red cross marks the desired teacher command and the mapped learner command.}
        \label{fig:commandPairs}
    \end{figure}

    An illustrative example of a set of command pairs is color-coded and presented in Figure~\ref{fig:commandPairs}. However, it is impractical to learn all the command pairs across the command domain. The proposed transferring framework relies on conformal mapping to find the rest of the command pairs by leveraging a limited number of learned command pairs.

    \vspace{5pt}
    \noindent
    \textbf{\emph{Schwarz-Christoffel Mapping-based Control Transferring}}
    
    Given the command pairs, the transferring process can be viewed as a geometric distribution transformation within the command domain. The proposed framework takes three steps to derive the learner command as depicted in Figure~\ref{fig:commandPairs_map}.
    First, the learner uses the teacher's control policy to generate a control input which is the teacher's desired command as if the learner was the teacher. Then, depending on user preference, the learner has the flexibility to choose multiple command pairs from the teacher's side, forming a poly-region that encompasses the teacher's desired command. The corresponding region on the learner's command domain is automatically determined by the learner's commands associated with the same command pairs as the teacher's vertices. At last, the Schwarz-Christoffel Mapping (SCM) is employed to conformally map the region in the teacher's command domain onto the region in the learner's domain. This 
    mapping allows us to pinpoint the precise learner command capable of producing maneuvers akin to those executed by the teacher in response to the desired command.
    
\begin{figure}[h]
        \centering
            \subfigure[]{
                \includegraphics[width = 0.45\textwidth]{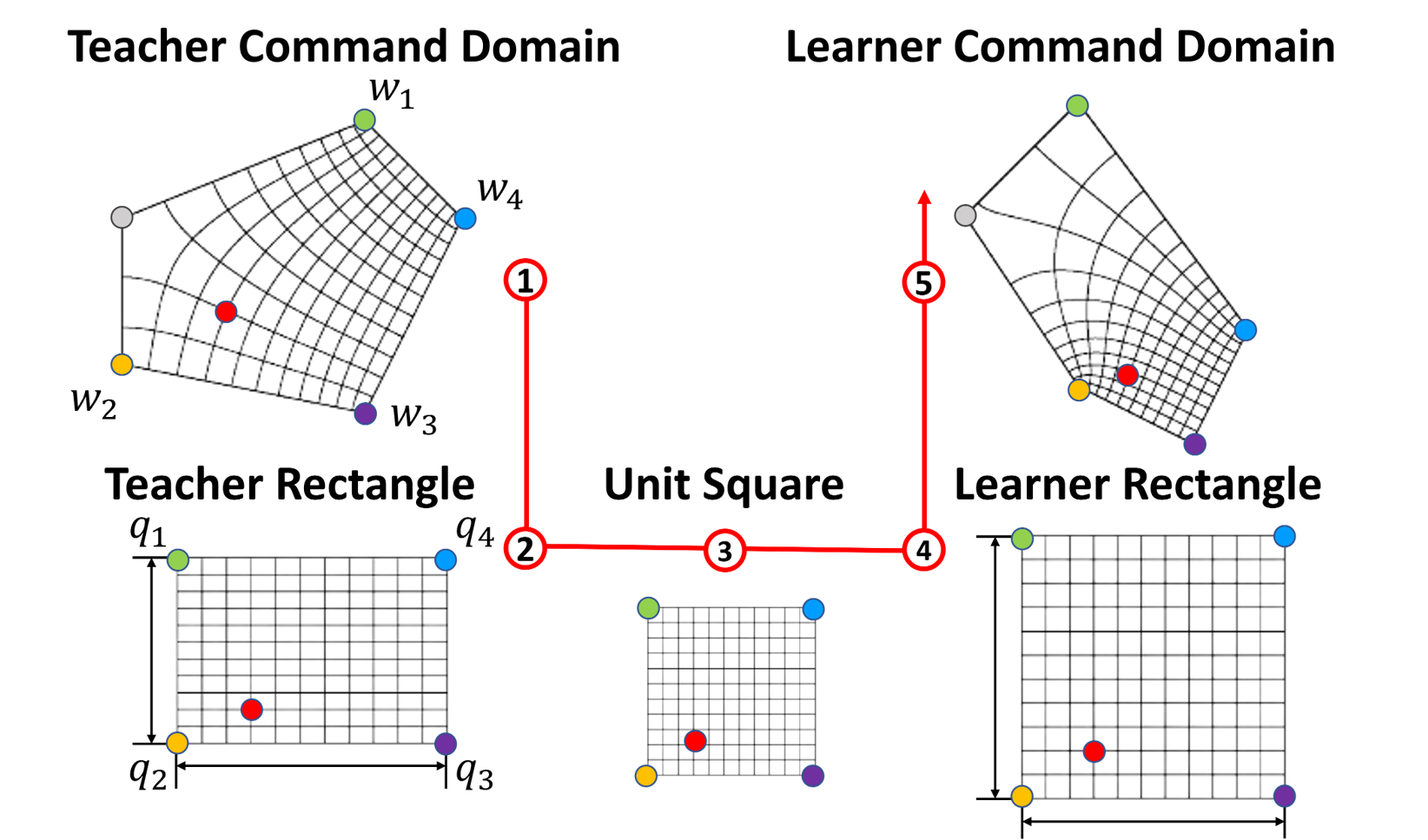}
                \label{fig:mappingFlow}
            }
            \hspace{5pt}
            \subfigure[]{
                \includegraphics[width =0.45\textwidth]{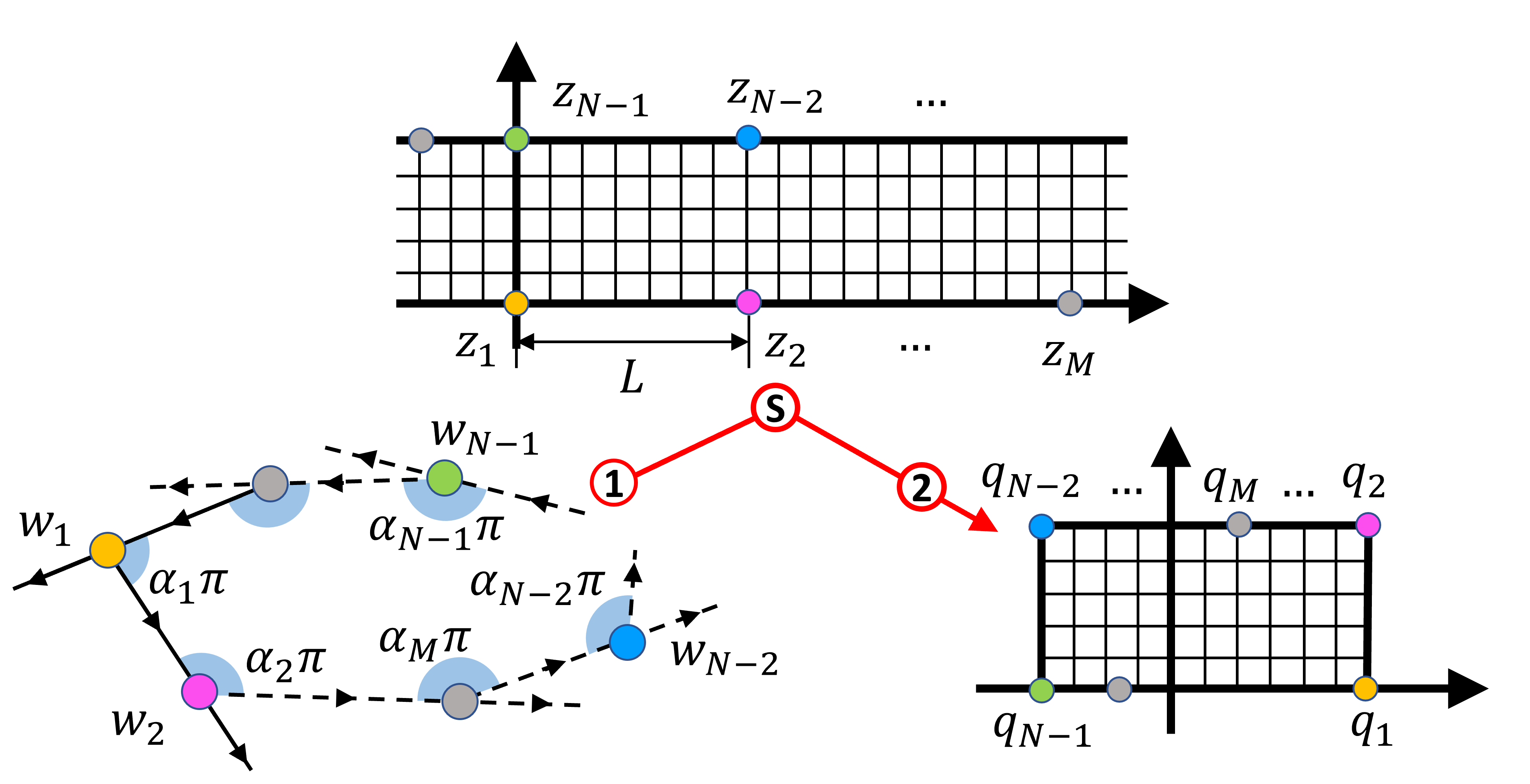}
                \label{fig:stripMapping}
            }
        \caption{(a) Mapping flow of transferring the desired teacher command to the learner; (b) Mapping of the polygon to a rectangle while using the bi-infinite strip as the intermediate plane.}
        \label{fig:SCMMapping}
    \end{figure}

    For ease of comprehension, we guide our readers through the mapping process using an example depicted in Figure~\ref{fig:SCMMapping}. In the context of this work, we employ a variant of the SCM theory which achieves a mapping from the interior of a polygon to a rectangle. The overall mapping flow of the proposed framework is shown in Figure~\ref{fig:mappingFlow} where polygons from both sides are first mapped to unique rectangles and a unit square is borrowed to bridge both ends. During the process, a bi-infinite strip is leveraged for using SCM to map a polygon to a rectangle (from 1 to 2, 5 to 4) as detailed in Figure~\ref{fig:stripMapping}. 

    Specifically, to compute the SCM function from either command domain to a rectangle, a polygon $\tau$ from the command domain is put onto the complex domain with the real axis representing the linear velocity and the image axis for the steering angle. Consider an irregular polygon $\tau$ with vertices denoted as $w_1, ..., w_N$ (where $N{\geq} 4$), arranged in a counterclockwise manner as illustrated in Figure~\ref{fig:stripMapping}. The interior angles at each vertex, $\alpha_1\pi, ..., \alpha_n\pi$, represent the angles formed by the edges originating from and terminating at that particular vertex. Initially, the prevertices $z_1, ..., z_N$ on the bi-infinite strip $\mathbb{S}$ are mapped to the vertices $w_1, ..., w_N$ on the polygon $\tau$ through:
    \begin{equation} \label{eql:strip2polygon}
        w =  f_{\mathbb{S}}^{\Gamma}(z) = A\int_{0}^{z}\prod_{j=0}^{N} f_{j}(z)dz +C
    \end{equation}
    where $A$ and $C$ are complex constants that rotate, translate, and scale the polygon and are determined by the shape and location of $\mathbb{P}$. Each factor $f_j$ sends a point on the boundary of the strip to a corner of the polygon while preserving its interior angles. By leveraging the Jacobi elliptic of the first kind~\cite{Byrd1971Handbook, yufeng2019schwarz}, the SCM mapping from the bi-infinite strip $\mathbb{S}$ to the rectangle $\mathbb{Q}$ is defined by:
    \begin{equation} \label{eql:Jacobielliptic}
        z = f_{\mathbb{Q}}^{\mathbb{S}}(q) = \frac{1}{\pi}\cdot \ln(\sin(q|m))
    \end{equation}
    where $q$ is the point on a regular rectangle and $m$ is the modulus of the Jacobi elliptic that is decided by $q$. The details of this conformal mapping can be found in~\cite{driscoll2002schwarz}. With Eqs. \eqref{eql:strip2polygon} and \eqref{eql:Jacobielliptic}, a desired teacher command located in the $\tau$ is conformally mapped to a regular rectangle by:
    \begin{equation}
        q = f_{SCM}(w) = {f_{\mathbb{Q}}^{\mathbb{S}}}^{-1}({f_{\mathbb{S}}^{\Gamma}}^{-1}(w)).
    \end{equation}
    The mapping function on the learner side is constructed following the same fashion. A unit square is borrowed to bridge between the two mapped rectangles resulting in a complete mapping process from teacher to the learner, such that any teacher command that falls in the teacher's mapping area is connected to an image on the learner side. 
    
    \subsection{Path Planner Transfer}
    The motion limit of the learner is associated with the commands residing on the boundary of the learner's command domain. To characterize the learner's limits within the teacher's command domain, we leverage command pairs with the learner component on its boundaries. The equivalent teacher commands from these pairs form a convex hull that effectively indicates the learner's limits. Commands within the interior of the convex hull correspond to feasible motions for the learner. Figure~\ref{fig:capability_discrete} demonstrates this method. Establishing such a convex hull provides evidence for imposing constraints on the teacher's controller and the path planner, ensuring that the commands generated remain within the hull, thus making the path achievable and able to be accurately followed by the learner. In cases where the command pairs used for constructing the convex hull do not confine to the learner's boundary, the corresponding area of the convex hull does not encompass the entire learner's command domain, suggesting that the transfer is not fully exploiting the learner's potential. Conservatively, a straightforward approach is to assess the learner's limit before transferring, isolating it from the transfer process through a dedicated calibration stage. During this calibration phase, a series of extreme learner commands are imparted to the learner to assess its motion limits. Upon the conclusion of the calibration stage, the characterized learner's limit is finalized and is used for imposing the constraints on the teacher's controller and planner. 

\begin{figure}[h]
            \subfigure[]{
            \includegraphics[width = 0.50\textwidth]{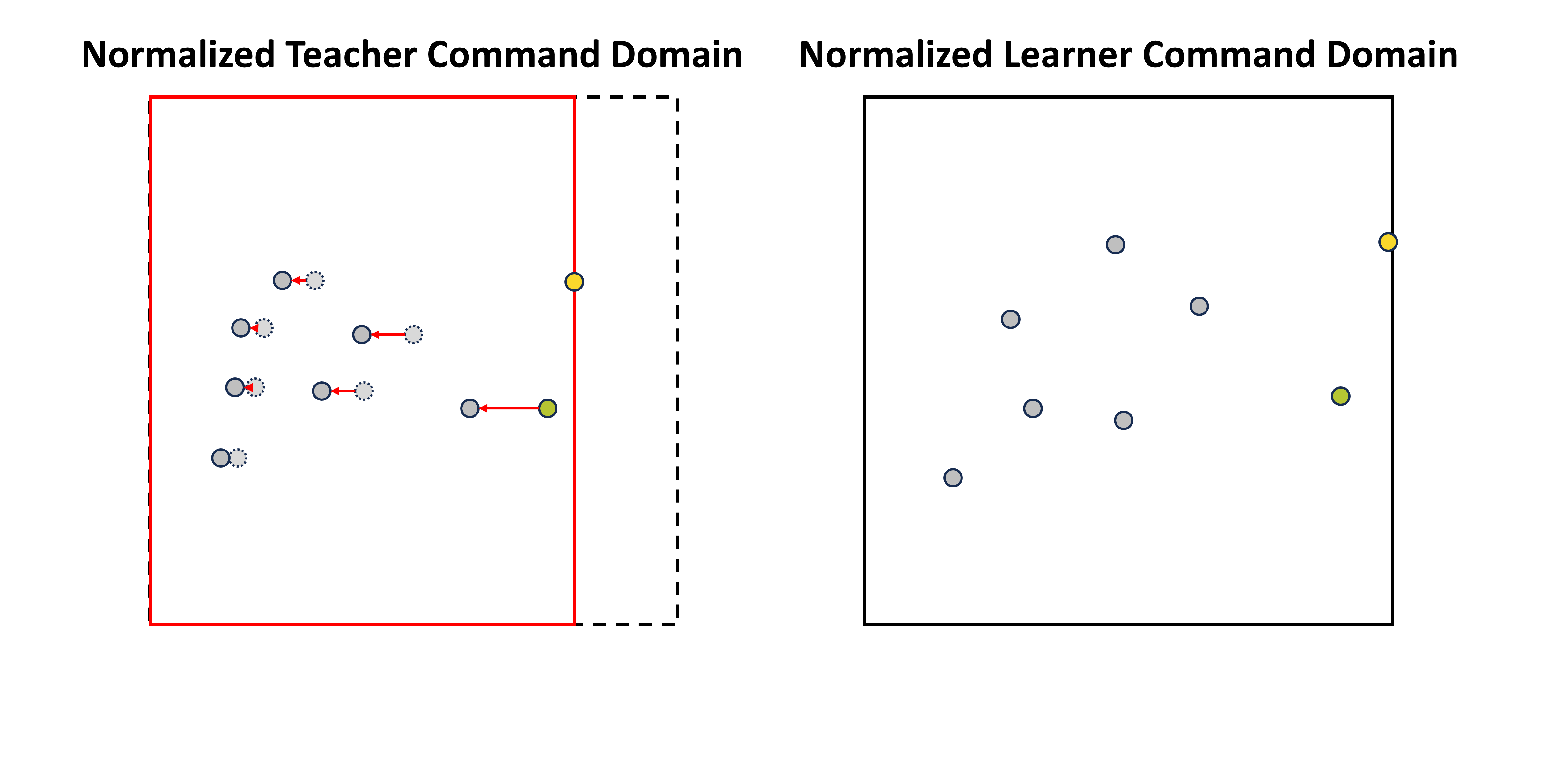}
            \label{fig:learn_capability_onEdge}
            }%
            \subfigure[]{
            \includegraphics[width =0.50\textwidth]{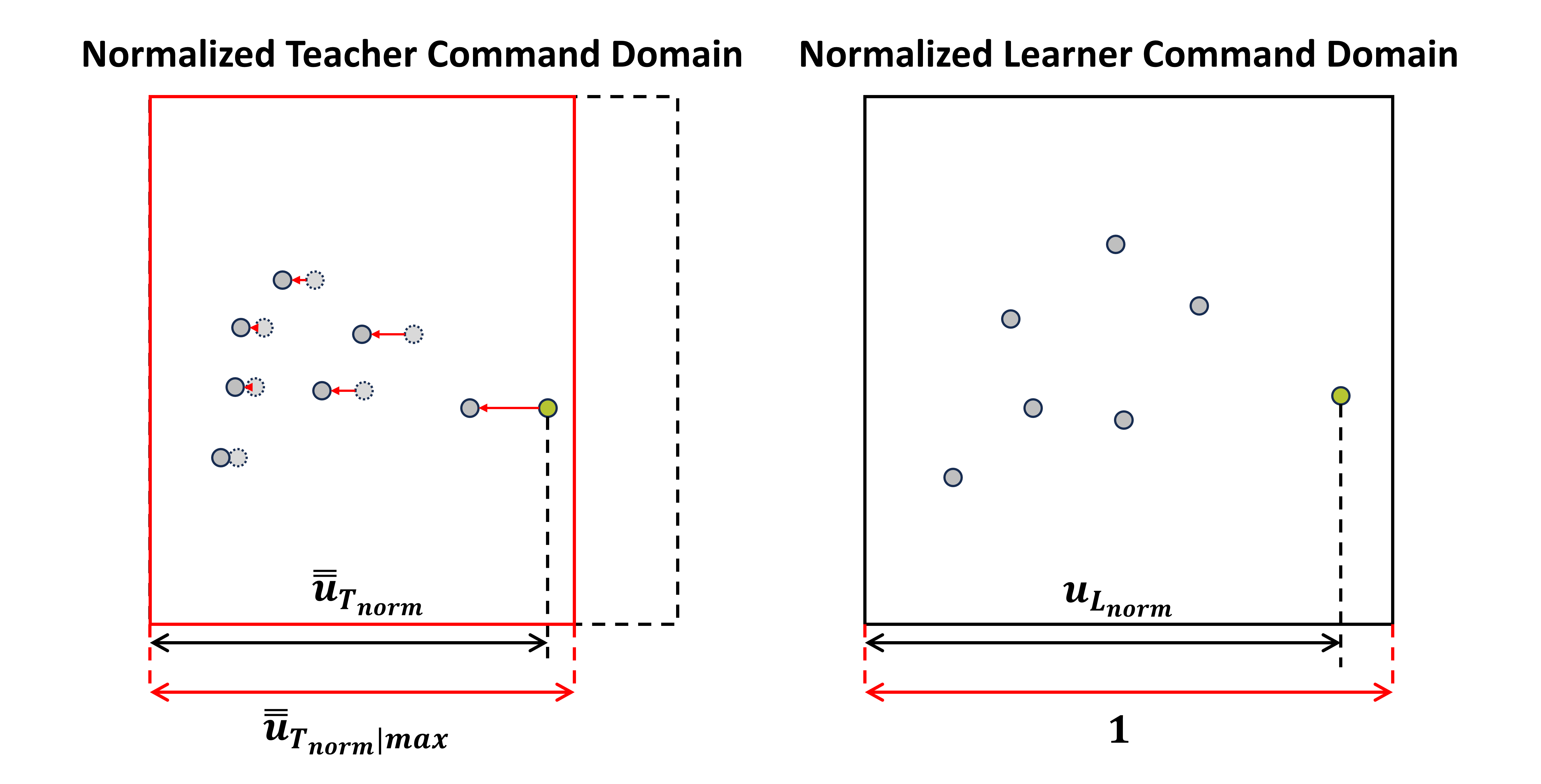}
            \label{fig:learn_capability_proportion}%
            }%
        \caption{Examples of characterization of learner's limits on the teacher command domain. (a) the new command pair directly marks the boundary of allowable teacher command; (b) proportionally shrinks the teacher's command space to approximate the learner's limits as the learner portion of the command pair is an outlier but not on the boundary.}
        \label{fig:learn_capability}
        \end{figure}

    Alternatively, another approach is to dynamically adapt the boundary of the learner's limit whenever it is needed along with the transferring process. When a new command pair, which behaves as an outlier in comparison to all existing command pairs, is introduced, it triggers a recalculation of the boundary in the teacher's command domain. On the one hand, if the newly established command pair resides on the boundary of the learner's command domain, the equivalent teacher command marks the upper or lower boundary of allowable teacher commands. Subsequently, all other existing normalized learner command pairs are also re-scaled to account for the adjustments in the command range. Figure~\ref{fig:learn_capability_onEdge} presents an example for this case, where the yellow dots indicate the newly added command pair. On the other hand, if the new command pair outlier does not align with the boundary of the learner's command domain, the range of permissible teacher commands undergoes proportional reduction. This proportional adjustment can be computed using Eq.~\eqref{eq:learn_capability}
     and is visually demonstrated in Figure~\ref{fig:learn_capability_proportion}.

        \begin{equation}
            \overline{\overline{u}}_{T_{norm} | max} = \frac{\overline{\overline{u}}_{T_{norm}}}{u_{L_{norm}}}
            \label{eq:learn_capability}
        \end{equation}

\section{Case Studies} \label{sec:case_study}
To show the generality and effectiveness of the proposed transferring framework, in this section, we showcase transferring two types of representative control and path planning methods. In the first study, we incorporate our framework with a motion primitive-based planning approach, one of the widely used searching-based approaches in robotics \cite{dharmadhikari2020motion, mueller2015computationally,fox1997dynamic}. The command sequence associated with the selected motion primitive is used for transferring and controlling the learner. Given the limited number of primitives, this study transfers the control inputs in a discretized command space. Such planner can benefit from a smaller discrete optimal searching space. In the second case, we transfer a Model Predictive Controller (MPC), a commonly used control strategy~\cite{li2015trajectory,sun2017disturbance,worthmann2015model}, from the teacher to control the learner over a continuous command space. The learner can benefit from the continuous control input for a smoother and closer tracking performance. Both scenarios operate effectively whether or not the learner's limits are pre-configured.

    \subsection{Motion Primitive-Based Transferring}\label{subsec:discrete_transfer}
    The overall structure of the approach is depicted in Figure~\ref{fig:overallFramework}. A motion primitive is a sequence of states that a vehicle performs within a short period. Motion primitives-based path planning utilizes a library of predefined motion primitives to search for and compose a sequence that closely approximates the desired path within certain bounds. In this case study, the learner's limits are determined in advance by collecting command pairs $\bm{u_p} {=} \langle\bm{u_{T}}, \bm{u_{L}}\rangle $ where $\bm{u_{L}}\in \{\bm{\underline{u}_L} \cup \bm{\overline{u}_L}\}$. Figure~\ref{fig:PrimitivePathPlanning} uses an example to show the command pairs in gray dots and highlights the learner's limits with the white envelope. Each teacher motion primitive $p_i {=} [ {\bm{x_T}}_1,{\bm{x_T}}_2,\dots,{\bm{x_T}}_t ]$ is created by applying a fixed control input to the teacher for a specified duration. The primitives and their corresponding commands are color-coded in the figure. The transfer of the path planner is achieved by excluding primitives whose associated control inputs exceed the boundary of the learner's limit, as illustrated by the crossed-out motion primitives and their corresponding control inputs. During the planning phase, the teacher's path planner searches primitives from the library and evaluates the difference between each of the primitives and the corresponding segment on the desired path $P$. As shown in Eq.~\eqref{eql:differencePrimitivePathPlanning} and in Figure~\ref{fig:PrimitivePathPlanning}, the difference is measured by considering both the dynamic time warping (DTW) distance $e_{d}$ between the motion primitive and the corresponding segment of the desired path and the heading difference $e_{\theta}$ at the end of the primitive:    
    

    \begin{figure}[h]
        \centering
        \begin{minipage}[t]{0.4\textwidth}
            \centering
            \includegraphics[height=3.2cm,keepaspectratio]{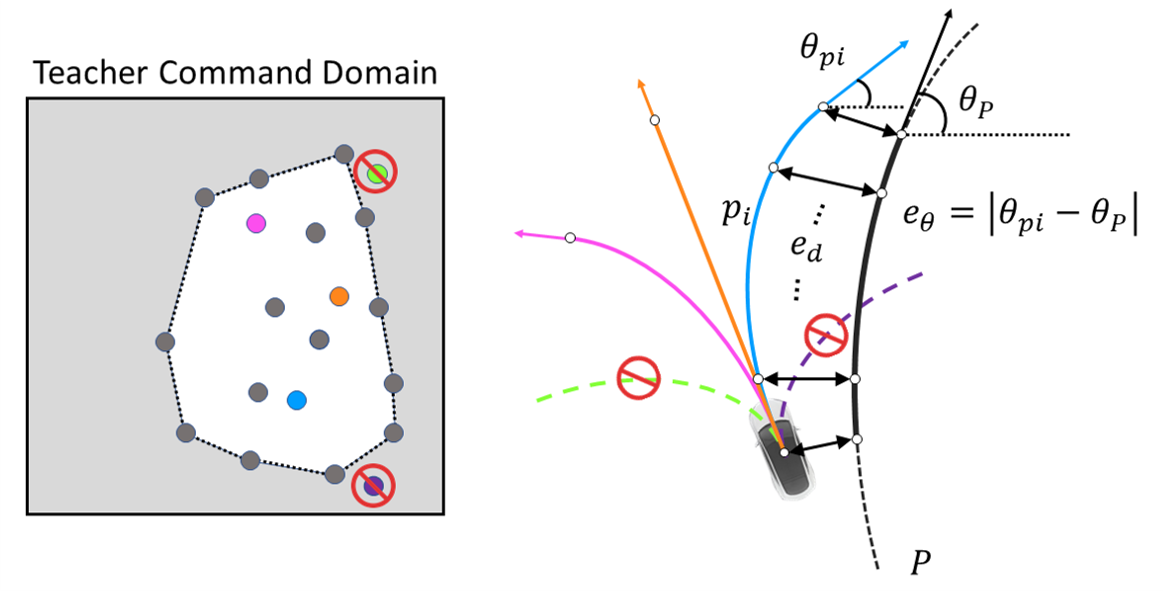}
            \caption{An example of the transfer framework with primitive-based path planning.}
            \label{fig:PrimitivePathPlanning}
        \end{minipage}%
        \hfill
        \begin{minipage}[t]{0.6\textwidth}
            \centering
            \vspace{-60pt} 
            \begin{equation}
            \begin{split}
                \delta_i&= k_d \cdot e_{d} + k_{\theta} \cdot e_{\theta} \\
                        &= k_d \cdot DTW(P, p_i) + k_\theta \cdot | (\theta_{P} - \theta_{p_i})|\\    
                p_{i}^* &= \min_{p_1, ..., p_i} \delta_i.
            \end{split}
            \label{eql:differencePrimitivePathPlanning}
            \end{equation}
        \end{minipage}
    \end{figure}

    
    The two types of differences are adjusted by user-defined gains ($k_d {\geq} 0$, $k_\theta {\geq} 0$). A higher $k_d$ ensures that the vehicle remains closer to the trajectory, while a higher $k_\theta$ increases the likelihood of selecting primitives that align more closely with the trajectory's orientation. Using these metrics, the planner evaluates all the primitives in the library, selecting the one that minimizes deviation as the optimal local path plan $p_i^{}$. The associated teacher's control input $\bm{u_{T}}^{}$ is then the command that will be transferred to the learner. The process of selecting command pairs and constructing the mapping polygon region can be outlined in the following steps:
    \begin{enumerate}
        \item \textbf{Perform Delaunay Triangulation:} Execute Delaunay triangulation on the teacher command domain to efficiently identify the triangle that includes $\bm{u_{T}}^{}$ inside.
        \item \textbf{Construct Teacher Polygon:} Combine the triangle that covers $\bm{u_{T}}^{}$ with one of its adjacent triangles to obtain the four vertices needed for constructing the polygon on the teacher's side.
        \item \textbf{Determine Learner Polygon:} Follow the same command pairs, the corresponding polygon on the learner's side is determined.
        \item \textbf{Verify Polygon Simplicity:} Ensure that both resulting quadrilaterals qualify as simple polygons by verifying that there are no edge crossings, except at the vertices.
    \end{enumerate}
   If edge crossings are detected during the simplicity check in Step 4, a different adjacent triangle should be selected in Step 2 to re-select the vertices and adjust the polygon formation. These steps ensure that the polygons are correctly formed and meet the required geometric conditions for computing SCM functions.



    Due to differences between the teacher and the learner, the learner may not perfectly replicate the teacher's motion, particularly when there are insufficient command pairs near $\bm{u_{T}}$, forcing the use of command pairs that are farther from it to construct the mapping region in Step 2.
    These motion deviations are typically not critical since the command sequence only lasts a short period and the planner can correct it at the next planning step. However, such deviation can pose safety risks when the learner operates in a cluttered environment. To enhance safety, we implement an event-triggered mechanism that continuously monitors the learner during operation. The monitor measures the distance $d_{\hat{e}}$ between the learner and the planned path, triggering a re-planning procedure if the deviation exceeds a threshold $\epsilon$. The threshold is designed to be dynamically changed and correlated with the minimum distance between the vehicle and the surrounding obstacles in the vehicle's field of view (FOV). Specifically, the threshold is defined as:
    \begin{equation}
        \epsilon =
        \begin{cases}
        \eta \cdot \min(|| \bm{x_L} -  \bm{x}_{\mathcal{O}i}||) & i = 1,2, \ldots , N_{\mathcal{O}}, \\
        \overline{\epsilon} & i = \varnothing, \\
        \end{cases}
    \end{equation}
    where $N_{\mathcal{O}}$ is the number of obstacles in the learner's field of view, $\bm{x}_{\mathcal{O}i}$ is the position of obstacle $i$, $\overline{\epsilon}$ denotes the maximum deviation allowed, and $\eta$ is a constant. This setup enables the planning process to adopt a more conservative approach when navigating near obstacles, allowing for timely intervention before the learner completes the current local plan.
    
    \vspace{5pt}
    \noindent
    \textbf{\emph{Simulation Results}}\label{subsubsec:primitive_result_sim}
    
\begin{figure*}[t]
      \centering
      \includegraphics[width=\textwidth]{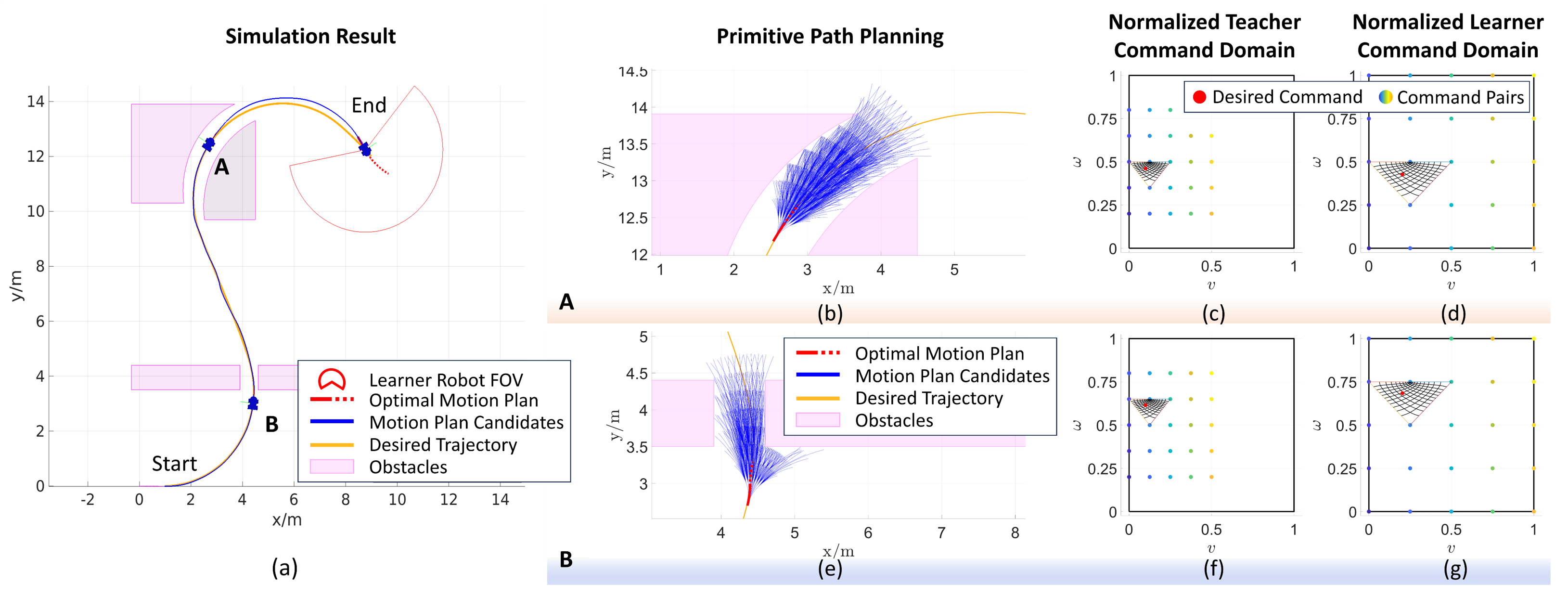}
      \vspace{-15pt}
      \caption{Simulation result of a learner following an ``S"-shaped path (a). The local path planning and the SCM mapping results for the robot at position `\textbf{A}' are shown in (b), (c), (d), and the results at position `\textbf{B}' are shown in (e), (f), and (g).}
      \label{fig:simulation_result}
    \end{figure*}

    We validate our transferring framework in simulation through a case where the vehicle suffers from compromised dynamics. The kinematics for the vehicle are given by the following unicycle model:

    \begin{align}\label{eql:motion_primitives_dynamicmodel}
    \begin{split}
        \begin{bmatrix} 
          x_{k+1} \\
          y_{k+1} \\
          \theta_{k+1} 
          \end{bmatrix} = 
          \begin{bmatrix}
              x_k\\
              y_k\\
              \theta_k
          \end{bmatrix} + \Delta t
          \begin{bmatrix} 
          v_k\cos\theta_k \\
          v_k\sin\theta_k \\
          \gamma_k
          \end{bmatrix}, \;\;\;
          \bm{u}_k &= \begin{bmatrix}
          v_k\\
          \gamma_k\\
          \end{bmatrix}
    \end{split}
    \end{align}
    
     where $v$ and $\omega$ denote the linear and angular velocities respectively. 
     The command ranges for both the teacher and the learner in this study case are detailed in Table~\ref{tb:motion_primitives_settings}. 
     The SCM method is implemented using the MATLAB Schwarz-Christoffel toolbox~\cite{driscoll2005algorithm} and a Gaussian noise of $G \sim \mathcal{N}(0, 0.1)$ is added to the learner's position to simulate measurement errors. The learner is asked to follow a ``S"-shaped trajectory through a cluttered environment as shown in Figure~\ref{fig:simulation_result}
     \begin{table}[h]
        \caption{Parameters for Motion Primitive Transferring Simulations and Experiments}\label{tb:motion_primitives_settings}
        \begin{tabular}{cc|cccc}
             &  & \multicolumn{1}{c|}{$\underline{v}~(m/s)$} & \multicolumn{1}{c|}{$\overline{v}~(m/s)$} & \multicolumn{1}{c|}{$\underline{\omega}~(rad/s)$} & $\overline{\omega}~(rad/s)$ \\ \hline
            \multicolumn{1}{c|}{\multirow{2}{*}{Simulation}} & Teacher & $0.0$ & $3$ & $-\pi/3$ & $\pi/3$ \\
            \multicolumn{1}{c|}{} & Learner & $0$ & $1$ & $-\pi/8$ & $\pi/8$ \\ \hline
            \multicolumn{1}{c|}{\multirow{2}{*}{Experiment}} & Teacher & $0$ & $1.6$ & $-1.2$ & $1.2$ \\
            \multicolumn{1}{c|}{} & Learner & $0$ & $1$ & $-1$ & $1$
        \end{tabular}
    \end{table}

\begin{figure*}[h]
      \centering
      \includegraphics[width=.9\textwidth]{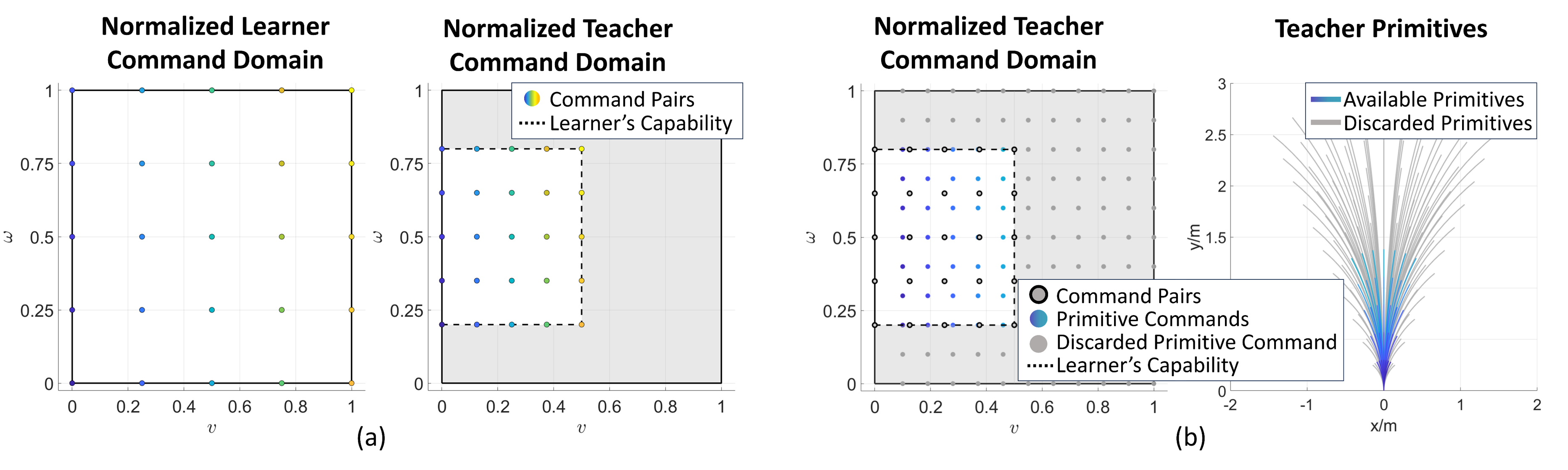}
      \vspace{-10pt}
      \caption{Examples of learner's limits assessment and transfer of path planner. (a) The command pairs are color-coded across the two command domains; (b)The primitives within the learner's limit are preserved for path planning.}
      \vspace{-10pt}
      \label{fig:learn_capability_sim_primitive}
    \end{figure*}

    As shown in Figure~\ref{fig:learn_capability_sim_primitive}(a), a $5\times5$ grid of command pairs are collected beforehand. The boundary of the command pairs on the teacher's command domain marks the limitation of the learner. Figure~\ref{fig:learn_capability_sim_primitive}(b) shows all of the teacher's motion primitives alongside their corresponding commands, each generated by driving the teacher with a specific control input for $1s$. Out of the 121 motion primitives, 35 are retained for path planning after excluding those that exceed the learner's limits.

    Figure~\ref{fig:simulation_result} shows two snapshots from the simulation, illustrating that the learner closely follows the desired trajectory. For the teacher's path planner, we set the planning horizon at $2$, meaning the local path consists of two primitives. The learner adopts more conservative maneuvers in clustered areas compared to open spaces, due to a smaller tolerance for deviation and more frequent triggering of replanning to enhance safe operations.

\begin{figure}[h]
        \centering
        \begin{center}
            \includegraphics[width=\textwidth]{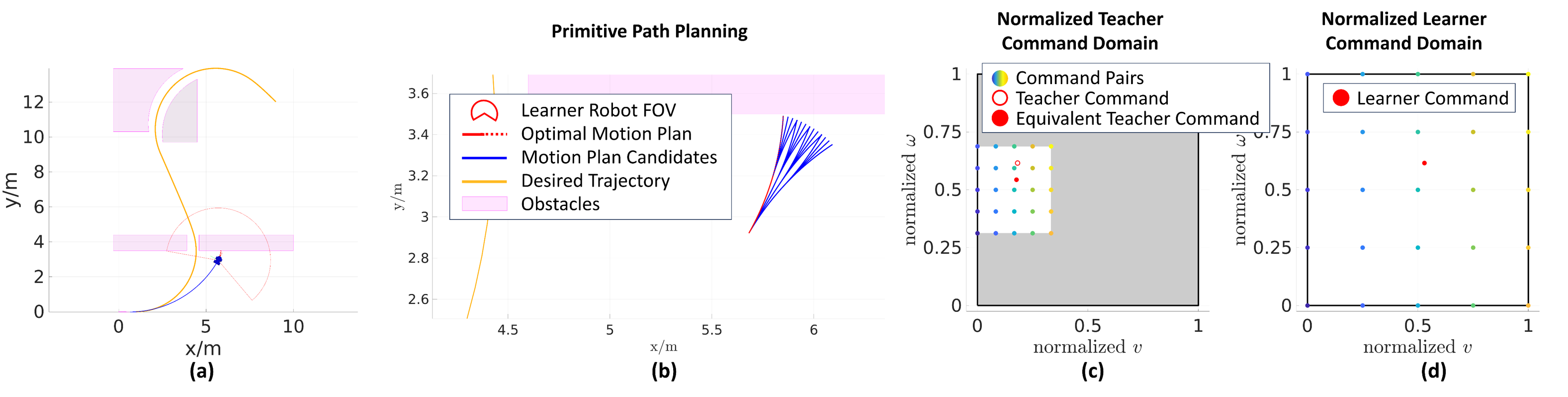}
        \end{center}
        \caption{Simulation results for baseline learner. (b),(c), and (d) show the primitive path plan, the normalized teacher command domain, and the normalized learner command domain at the time instance in (a).}
        \label{fig:simularion_noSCM}
    \end{figure}

    In Figure~\ref{fig:simularion_noSCM}, we show the result of the baseline learner for the same task. The baseline learner directly applies the teacher's commands, unaware of any necessary adjustments for its dynamics. As expected, the learner fails because it uses commands not adapted to its new dynamics.

    \newpage
    \vspace{5pt}
    \noindent
    \textbf{\emph{Experiment Results}}\label{subsubsec:primitive_result_exp}
    
    The real experiments, similar to the setup in simulation, are conducted for transferring the planning and control knowledge from the same simulated teacher to different learner vehicles in the real world. The command ranges for the vehicles are listed in Table~\ref{tb:motion_primitives_settings}. 
    The vehicle control is managed via the MATLAB ROS Toolbox in conjunction with \REV{the Robot Operating System (ROS)\cite{quigley2009ros}}. The experiments are conducted indoors, with the vehicles' states tracked by a VICON motion capture system. \REV{All physical experiments were conducted using different ground vehicles and a base computer equipped with an Intel i7-6500U and 8GB of RAM. For each experiment, the robot’s controller, planner, and the proposed transfer framework are run on the base computer and the transferred commands are directly transmitted to the robot through ROS. Because the MATLAB SCM Toolbox asymptotically approximates the solution to the SCM problem, the solving time varies depending on the shape of the mapping area. In our implementation, the entire pipeline typically runs at around 25 Hz (though this may fluctuate), while the experiments included in this work were conducted at 10 Hz, which proved sufficient for controlling the various robots examined in this study. For applications requiring faster control rates, more optimized or compiled implementations can significantly improve computational speed over the current implementation which relies on the MATLAB SCM Toolbox.}

\begin{figure*}[h]
      \centering
      \includegraphics[width=.9\textwidth]{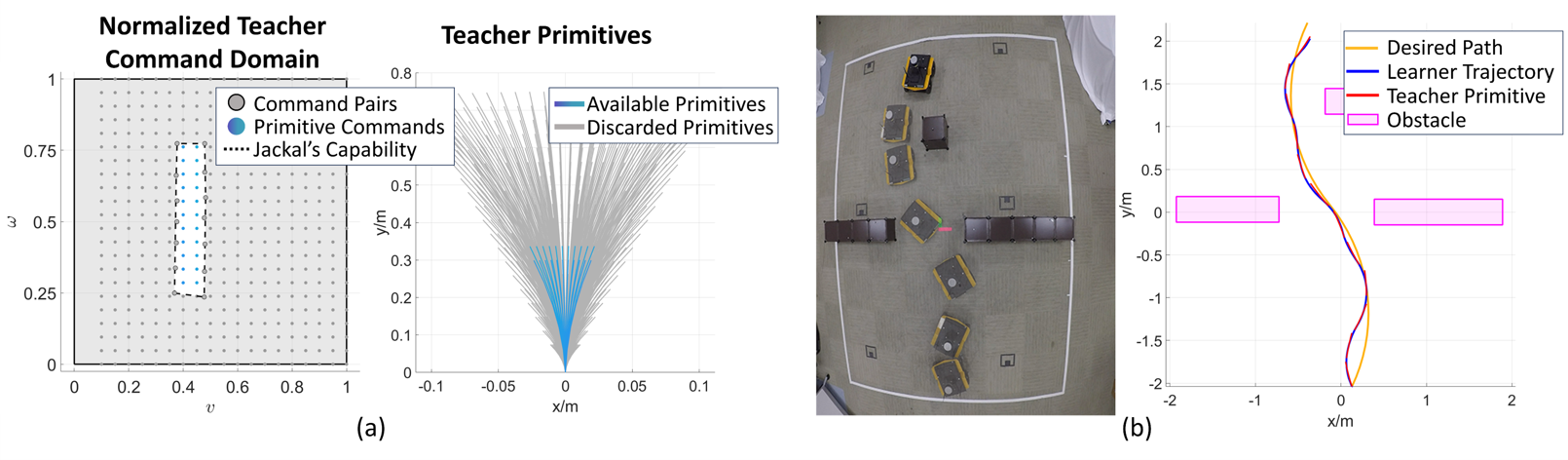}
      \caption{Figure (a) Jackal's limits is indicated within the white envelope. The gray points on the dashed boundary are the command pairs that calibrate the Jackal's limits. The blue-colored commands and primitives are used for the Jackal; (b) Jackal experiment with the proposed approach.}
      \label{fig:exp_result_primitive_jackal}
    \end{figure*}

\begin{figure}[h]
      \begin{center}
        \includegraphics[width=0.5\textwidth]{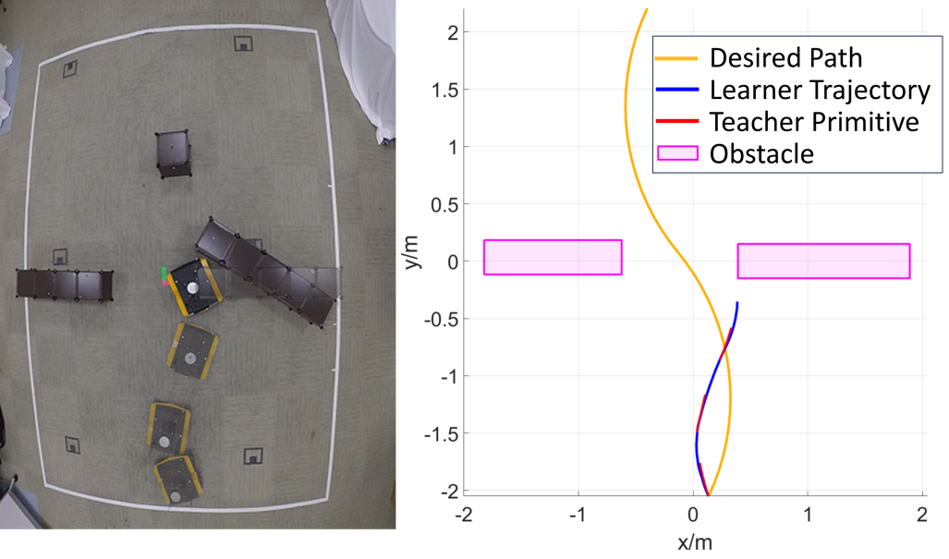}
      \end{center}
      \caption{Jackal experiment results for directly applying the teacher's commands.}
      \label{fig:ExpJackalNoSCM}
    \end{figure}

    In the first experiment in Figure~\ref{fig:exp_result_primitive_jackal}(b), a Clearpath Robotics Jackal UGV is treated as the learner vehicle for tracking an ``S"-shaped path through a gate. We characterize the vehicle's limits by constructing initial command pairs with a set of commands that consists solely of its upper and lower boundary commands, each executed for a duration of $1s$.
    The command pairs and the teacher's primitives used for planning the learner's path are displayed in Figure~\ref{fig:exp_result_primitive_jackal}(a). To assess the robustness of our proposed approach, the learner's initial heading is set with a $\frac{\pi}{4}\textit{rad}$ offset from the desired orientation. During the tracking task, the maximum distance recorded between the desired path and the actual trajectory was $0.1905m$, while the maximum deviation between the actual trajectory and the local motion plan was $0.0293m$. Given the vehicle's initial misalignment with the desired path and its dimensions of approximately $0.5m$$\times0.43m$$\times0.25m$, this deviation can be considered negligible. For comparison, the baseline experiment was conducted without the SCM component and the results are shown in Figure~\ref{fig:ExpJackalNoSCM}.

\begin{figure*}[t]
      \centering
      \includegraphics[width=1\textwidth]{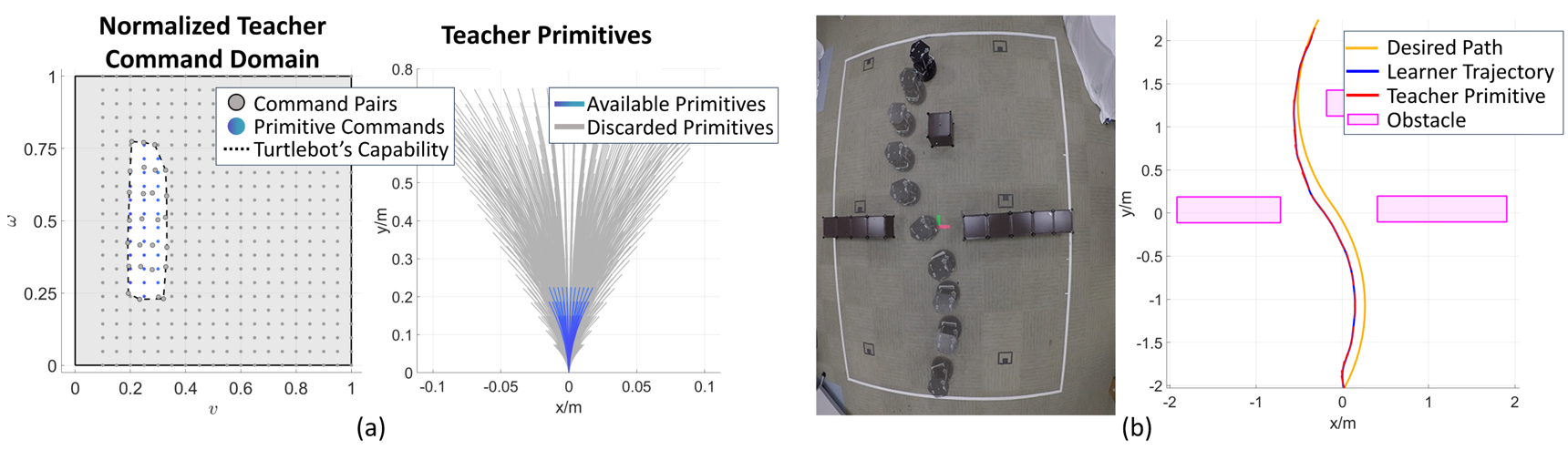}
      \vspace{-15pt}
      \caption{(a) Command pairs and primitives for the Turtlebot-2 experiment ; (b) Turtlebot-2 experiment snapshots with the proposed approach.}
      \label{fig:exp_result_primitive_turtlebot}
    \end{figure*}

    To demonstrate the generalizability of our proposed framework, we conducted an additional experiment using a Turtlebot-2 UGV as the learner with the same learner configuration listed in Table~\ref{tb:motion_primitives_settings}. The command pairs and primitives used for the Turtlebot are depicted in Figure~\ref{fig:exp_result_primitive_turtlebot}(a). The results indicate that, with our proposed approach, the Turtlebot successfully adapted the teacher's controller and path planner to follow the desired path with a maximum deviation of $0.1381$~m. The tracking error between the learner’s trajectory and the planned primitive remained within $0.0978$~m, demonstrating close alignment as illustrated by the nearly overlapping path plans and final trajectory in Figure~\ref{fig:exp_result_primitive_turtlebot}(b).

    \subsection{Continuous Space Transferring} \label{subsubsec:continuous_transfer}
    One of the primary limitations of motion primitive-based path planning transfer is that the commands available for mapping are constrained by the number of motion primitives in the library. Once the motion primitive is selected through an exhaustive search, the learner must map and execute the associated command sequence, either until completion of the current sequence or until an event-triggered re-planning intervenes. This results in the learner having only a limited number of discrete commands to choose from. Additionally, the fixed length of the command sequence may hinder the learner's ability to quickly adjust to deviations. Furthermore, significant effort is required to obtain the motion primitives and calibrate the learner's limits in advance. To address these limitations, we introduce a case study that incorporates a receding horizon controller within our proposed transfer framework. Specifically, we utilize MPC as a unified method for both control and local motion planning for the teacher. Additionally, we demonstrate how to dynamically learn and adjust the learner's limits from scratch without separating it from the transfer process.

    \vspace{5pt}
    \noindent
    \textbf{\emph{Model Predictive Controller}}\label{subsec:mpc}
    
    The simulated teacher uses the same model as introduced in Eq.\eqref{eql:motion_primitives_dynamicmodel}. The teacher exploits an MPC that tracks the desired path while avoiding the obstacles. The cost function and the optimal control problem are formalized as Eq.~\eqref{eq:MPC_OCP}.



    \begin{equation}
    \begin{aligned}
        & \ell(\bm{x}, \bm{u}) = {\lVert \bm{x} - \bm{x}^{\textit{ref}}  \rVert}_Q^2 + {\lVert \bm{u} - \bm{u}^{\textit{ref}}  \rVert}_R^2 \\
        & \min_{u}\quad    J_N(\bm{x}_0, \bm{u}) = \sum_{k = 0}^{N-1}\ell(\bm{x}_k, \bm{u}_k) \\
        & \quad \text{s.t.} \quad       \bm{x}_{k+1} = f_T(\bm{x}_k, \bm{u}_k)\\
        & \quad \quad \quad \lVert \bm{x}_k - \bm{x}_{\mathcal{O}i}  \rVert > r_i, \forall i \in [1,N_{\mathcal{O}}]\\
        & \quad \quad \quad \bm{x}_k \in X, \forall k \in [1, N]\\
        & \quad \quad \quad \bm{u}_k \in U, \forall k \in [1, N-1]\\
        & \quad \quad \quad \bm{u}' \leq \epsilon_u
    \end{aligned}
    \label{eq:MPC_OCP}
    \end{equation}
    where $N$ and $N_{\mathcal{O}}$ denote the predicting horizon and the number of obstacles respectively. $\bm{x}_{\mathcal{O}i}$ and $r_i$ denote the position and the radius of the $i^{th}$ obstacle. We cap the changing rate of the input at $\epsilon_u$ as we ignore the acceleration period out of simplicity. 
    \REV{$x$ and $u$ denote the system’s state and control input, respectively. The reference states $\bm{x}^{\textit{ref}}$ are sampled along the desired path, starting from the point closest to the teacher. Each subsequent point is spaced at the maximum distance the teacher can traverse along the path within one timestep, thus optimizing the time required to complete the task. The control references $u^{\textit{ref}}$ is initialized to zero during the first planning iteration and, in each subsequent iteration, is warm-started using the control sequence from the previous iteration, accelerating convergence to the optimal control inputs.} During the early stages of the transfer process, users have the flexibility to adjust the spacing between these reference states. This allows the teacher to perform a variety of maneuvers, which aids in populating the command pairs across the teacher's command domain. Whenever the command pairs are updated, leading to a refinement in the configured learner's limits on the teacher's command domain, the constraints on $U$ are revised to align with the updated operational boundaries of the learner. This adjustment ensures that the optimized control inputs consistently remain within the defined limits 
    of the learner.

    \vspace{5pt}
    \noindent
    \textbf{\emph{Command Pair Refinement}}\label{subsec:cmdPairs_cluster}
    
     When constructing command pairs, the process of identifying equivalent teacher commands is relatively straightforward in scenarios with subtle motion noise, as the same learner commands typically result in almost identical motions and, consequently, nearly identical equivalent teacher commands. However, when motion noises are non-negligible, variations in motions result in inconsistent equivalent teacher commands for even the same learner command, complicating the retrieval of the correct equivalent teacher command.
     
\begin{figure}[ht]
        \centering
        \begin{minipage}[t]{0.68\textwidth}
            \centering
            \includegraphics[width=\textwidth]{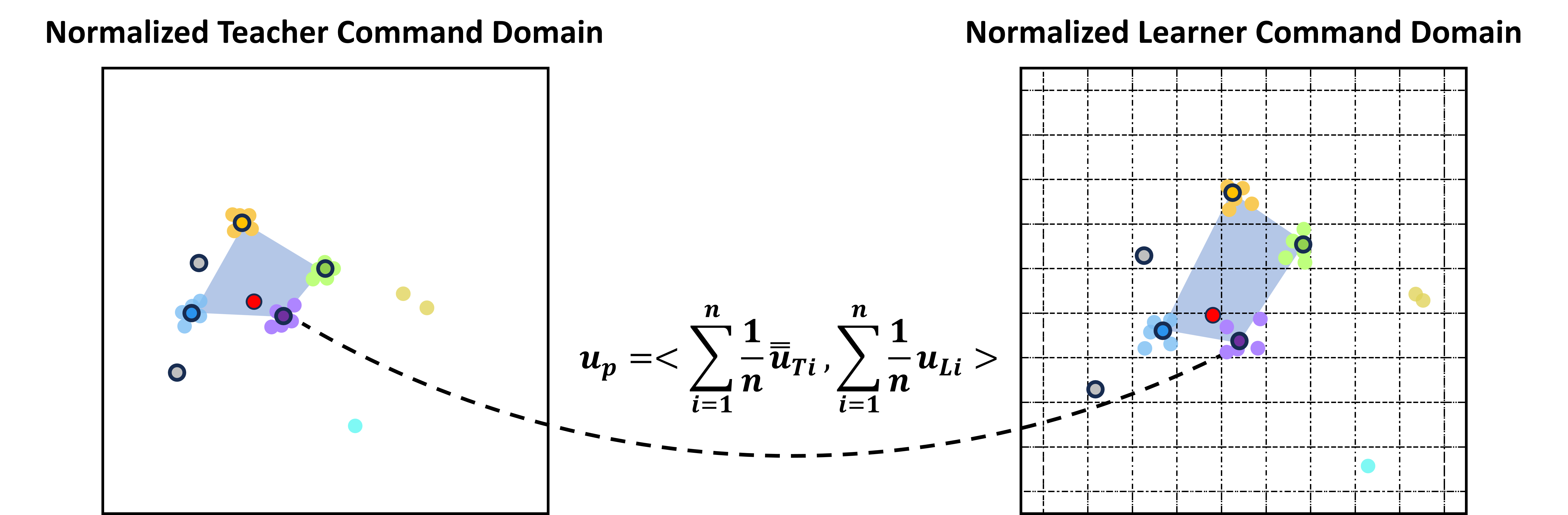}
            \caption{An example of the refined command pairs and the corresponding clusters. Command pairs are color-coded and shown in circles with black edges.}
            \label{fig:commandPairs_map_cells}
        \end{minipage}
        \hfill
        \begin{minipage}[t]{0.28\textwidth}
            \centering
            \includegraphics[width=\textwidth]{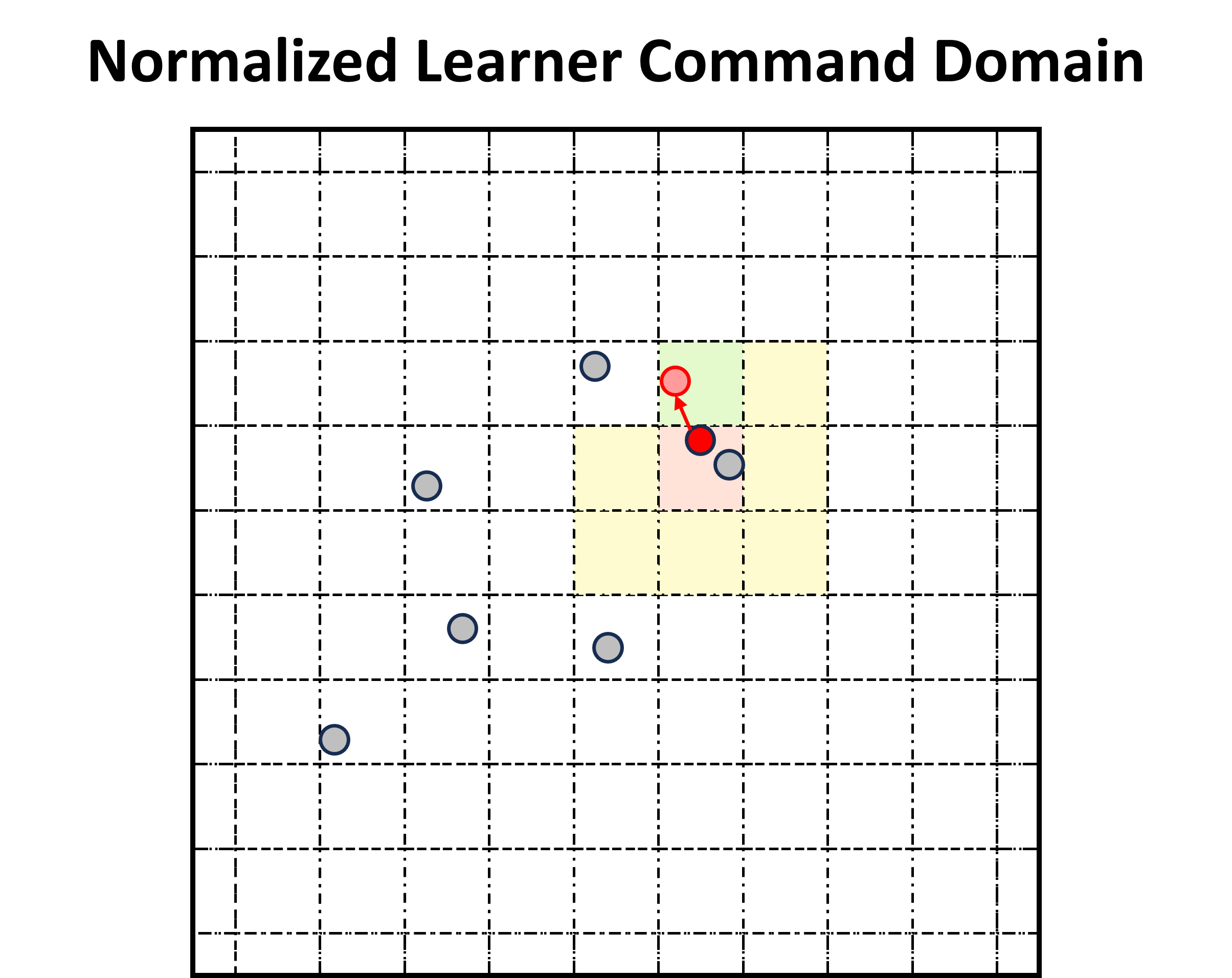}
            \caption{An example of perturbing the desired learner command to one of the adjacent empty command cells.}
            \label{fig:perturbateCommand}
        \end{minipage}
    \end{figure}

     To address this issue, we refine the definition of command pairs by first partitioning the normalized learner command domain into grid cells. Subsequently, the previously defined command pairs are grouped into clusters based on whether their learner components are within the same cell. 
    Each refined command pair is then derived by averaging those within the same cluster and a user-defined minimum cluster size, $k_{\text{min}}$, is introduced for constructing command pairs. 
     Figure~\ref{fig:commandPairs_map_cells} demonstrates the refined command pairs.


    \vspace{5pt}
    \noindent
    \textbf{\emph{Controller and Path Planner Transferring}}\label{subsec:scm_mpc_methods}
    
    If the learner's limits 
    have already been characterized, transferring the teacher's command to the learner resembles the previous case, except the available teacher commands are within the continuous space constrained by the learner's boundaries. In scenarios with few or no pre-learned command pairs, to facilitate accurate control transfer, it is desirable to construct as many command pairs as possible in a timely manner. For learners with non-negligible motion noise,  Increasing $k_{\text{min}}$ improves the precision of the command pairs but also prolongs the time required for constructing them. 
    To accelerate this process, without losing generality, we offer strategies to accelerate the construction of command pairs to compensate for the increased effort required in collecting more commands.

\begin{figure}[h]%
        \centering
        \includegraphics[width=0.95\textwidth]{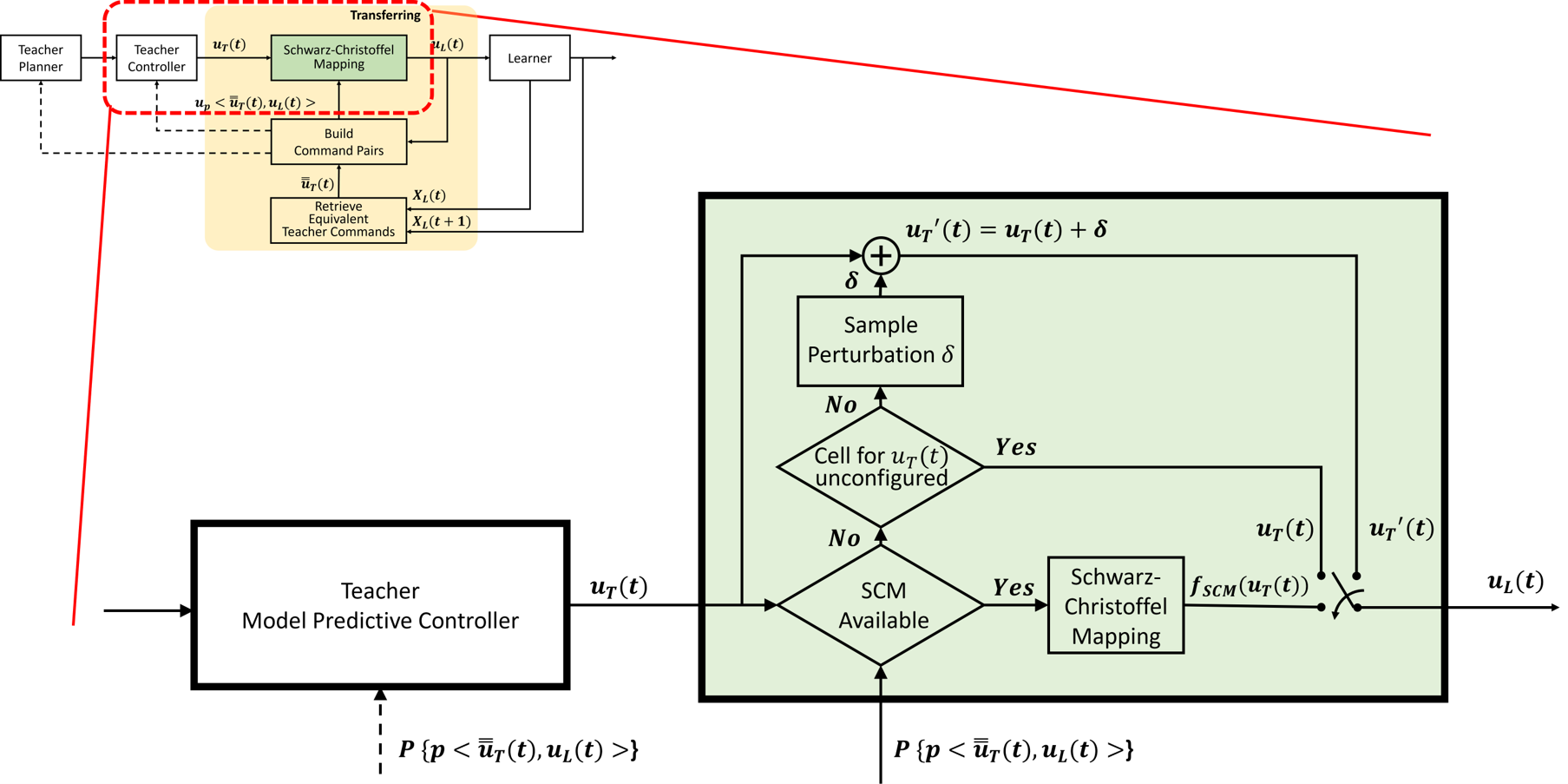}
        \caption{The block diagram of the proposed SCM-based learning framework without pre-learned learner's limits. \REV{The changes needed to accommodate unknown learner's limits are highlighted in the zoomed-in window.}}
        \label{fig:scm_mpc_overall}
    \end{figure}

    Our comprehensive framework with these enhancements is shown in Figure~\ref{fig:scm_mpc_overall}. We emphasize the components that help in scenarios where existing command pairs are insufficient. When existing command pairs cannot form a polygon area that includes the desired command, the cell containing the desired command may or may not already have a configured command pair. If no command pair is configured in the cell, the learner directly adopts the teacher's command, indicating that it has not yet recognized the local geometric differences across the command domains of the two systems. Conversely, if a command pair is present in the cell, the learner samples the desired command from unconfigured adjacent cells, as illustrated in Figure~\ref{fig:perturbateCommand}. In this example, existing command pairs are depicted as gray dots. Instead of directly using the teacher's command (located in the red cell), the learner selects a command from one of the adjacent cells.
    
    \vspace{5pt}
    \noindent
    \textbf{\emph{Simulation Results}}\label{subsec:mpc_scm_sim_results}
    
\begin{figure}[h]
        \centering
        \begin{minipage}{1.0\textwidth}
            \begin{equation}\label{eql:mpc_dynamicmodel_learner}
            \begin{bmatrix} 
                x_{k+1} \\
                y_{k+1} \\
                \theta_{k+1} 
            \end{bmatrix} = 
            \begin{bmatrix}
                x_k \\
                y_k \\
                \theta_k
            \end{bmatrix} + \Delta t
            \begin{bmatrix} 
                d(h_v(n(v_k)))\cos(\theta_k) \\
                d(h_v(n(v_k)))\sin(\theta_k) \\
                d(h_\omega(n(\omega_k)))
            \end{bmatrix}, 
            \bm{u}_k = \begin{bmatrix}
                v_k \\
                \omega_k
            \end{bmatrix}
        \end{equation}
        \end{minipage}
        \begin{minipage}{0.3\textwidth}
            \hfill
            \includegraphics[width=0.8\linewidth]{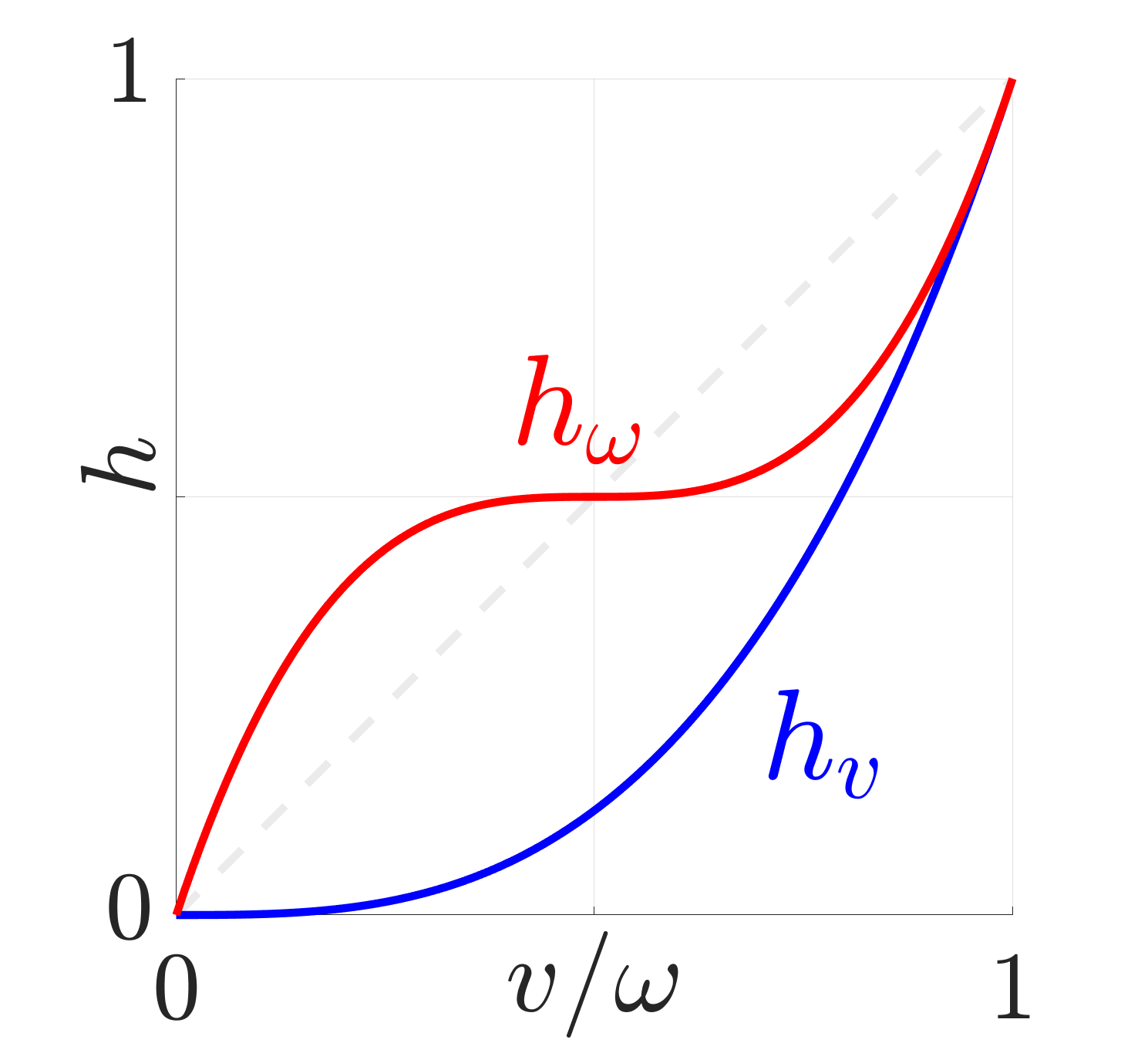}
            \caption{The nonlinear functions employed to alter the learner's dynamics in simulations.}
            \label{fig:nonlinear_sim}
        \end{minipage}%
        \begin{minipage}{0.7\textwidth}
            \begin{equation*}
                \begin{aligned}
                    \text{where,} \quad & h_v(v) = v^3 \hfill\\
                    & h_\omega(\omega) = 4\cdot(\omega-0.5)^3 + 0.5\\
                    & n(\times) = (\times - \underline{\times}_L)/(\overline{\times}_L - \underline{\times}_L)\\
                    & d(\times) = \times \cdot (\overline{\times}_L - \underline{\times}_L) + \underline{\times}_L\\ 
                \end{aligned}
            \end{equation*}
        \end{minipage}
    \end{figure}

    The initial command ranges for the teacher and the learner are outlined in Table~\ref{tb:scm_mpc_settings}. During the transferring process, the teacher's command range corresponds to the range of the refined learner's limits. We intentionally unbalanced the turning abilities towards both sides to increase the discrepancies between the two systems. The kinematic model of the simulated learner is similar to that of the teacher. However, to further demonstrate the effectiveness of the proposed approach, we introduced a nonlinear transformation within the learner's model to further increase the dynamical differences between the teacher and the learner system. 
    The specific formulation of the learner's model is described in~\eqref{eql:mpc_dynamicmodel_learner} and the employed nonlinear functions are depicted in Figure~\ref{fig:nonlinear_sim}. The learner's command domain is divided into an $11\times11$ grid for grouping and constructing command pairs, with a minimum cluster size of $(k_{min}=5)$ set for the simulation.

    \begin{table}[h]
        \caption{Parameters for MPC Transferring Simulations and Experiments}\label{tb:scm_mpc_settings}
        \begin{tabular}{cc|cccc}
             &  & \multicolumn{1}{c|}{$\underline{v}~(m/s)$} & \multicolumn{1}{c|}{$\overline{v}~(m/s)$} & \multicolumn{1}{c|}{$\underline{\omega}~(rad/s)$} & $\overline{\omega}~(rad/s)$ \\ \hline
            \multicolumn{1}{c|}{\multirow{2}{*}{Simulation}} & Teacher & $0.05$ & $0.6$ & $-\pi/4$ & $\pi/4$ \\
            \multicolumn{1}{c|}{} & Learner & $0.05$ & $0.3$ & $-\pi/16$ & $\pi/12$ \\ \hline
            \multicolumn{1}{c|}{\multirow{2}{*}{Experiment}} & Teacher & $0.05$ & $0.6$ & $-\pi/4$ & $\pi/4$ \\
            \multicolumn{1}{c|}{} & Learner & $0.05$ & $0.2$ & $-\pi/8$ & $\pi/8$
        \end{tabular}
    \end{table}
    
    \begin{figure*}
        \centering
        \includegraphics[width=\textwidth]{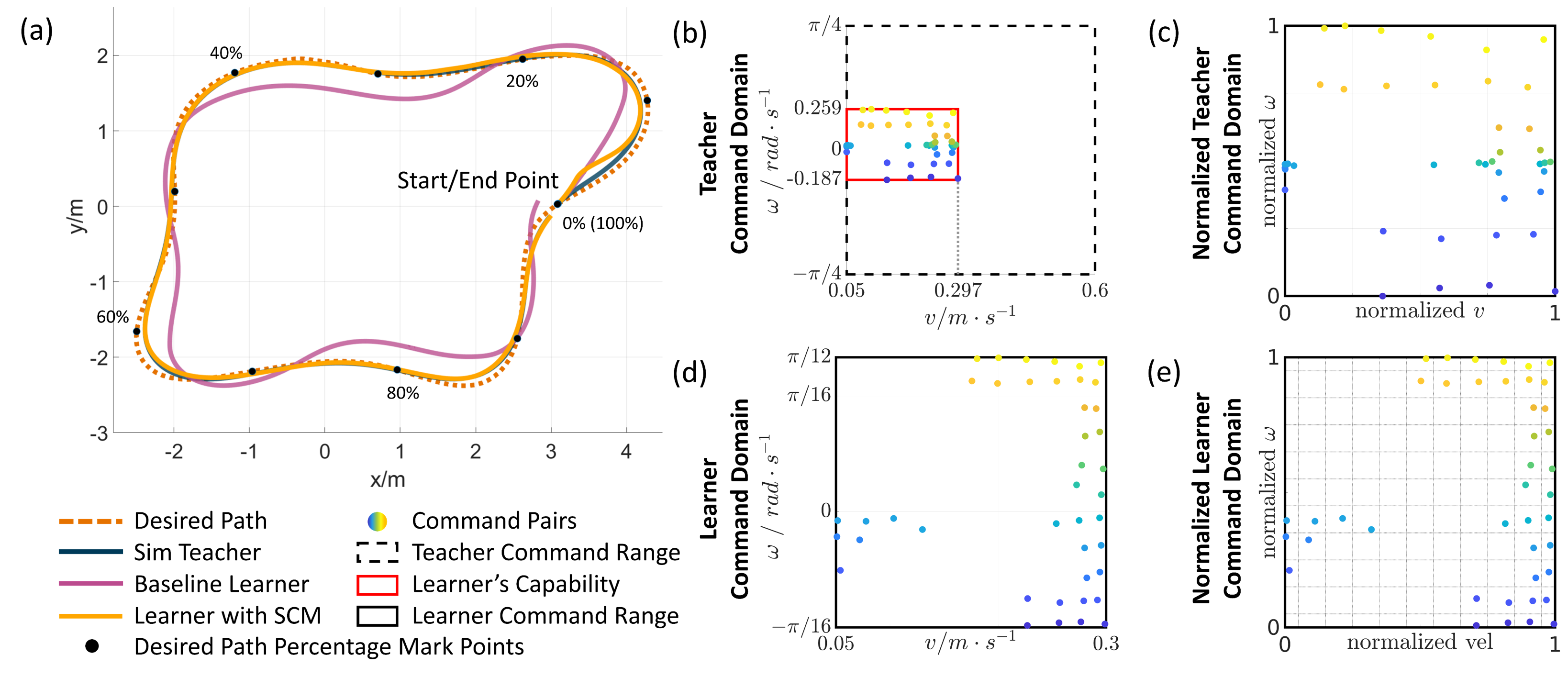}
        \caption{Simulation results of SCM-based transfer with MPC. (a) trajectories comparison; (b) teacher command domain; (c) normalized teacher command domain; (d) learner command domain; (e) normalized learner command domain.}
        \label{fig:SCM_MPC_sim_result}
    \end{figure*}
    
    Figure~\ref{fig:SCM_MPC_sim_result} illustrates the results of following a desired path without any pre-existing command pairs. In this setup, we compare the performance of a learner using our proposed transfer framework against an ideal teacher and another learner that directly applies the teacher's commands as the baseline
    . All three simulations start from the same initial pose, intentionally misaligned with the desired path. Figure~\ref{fig:SCM_MPC_sim_result}(a) 
    shows that using our approach, the learner's trajectory closely follows that of the ideal teacher, except during an initial adjustment period while the framework adapts to the learner's capabilities. The command pairs built during the tracking task are depicted in  Figure~\ref{fig:SCM_MPC_sim_result}(b) through (e). The learned learner's limits are marked by a red rectangle on the teacher's command domain which closely approximates the properties set in Table~\ref{tb:scm_mpc_settings}. The significant nonlinearity between the two systems is evident when comparing the distribution of command pairs across the command domains of both systems ( Figure~\ref{fig:SCM_MPC_sim_result}(b) and (d)).

    \begin{figure*}[t]
        \centering
        \includegraphics[width=1.0\textwidth]{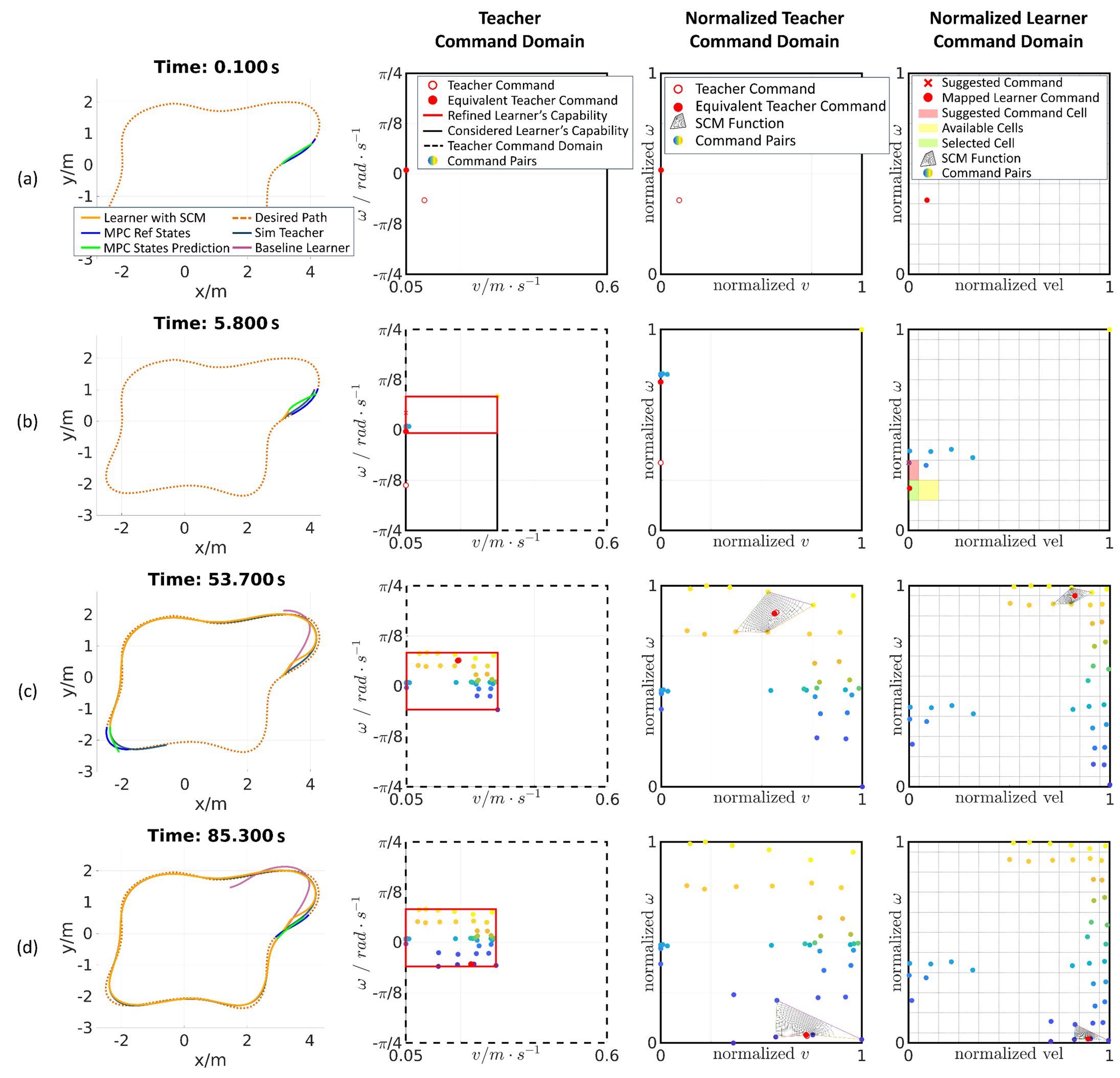}
        \caption{Examples from simulations demonstrating the use of SCM to transfer MPC from a teacher to a learner at various time frames. Each column: 1) compares the trajectories between the simulated teacher, baseline learner, and our learner with SCM; 2) teacher command domain and configured learner's limits; 3) normalized teacher command domain within the learner's boundary; 4) normalized learner's command domain.}
        \label{fig:SCM_MPC_sim}
    \end{figure*}
    
   
    Figure~\ref{fig:SCM_MPC_sim} presents snapshots at different stages of the simulation. The first column compares the trajectories and tracking progress at specified times. Row (b) highlights how the learner perturbed the original desired command to expedite the construction of command pairs in an unconfigured cell. This perturbation resulted in introducing a new command pair, leading to proportionally shrink the learner's minimum angular velocity limit, as indicated by the red rectangle on the teacher's command domain. Row (c) displays an example of using SCM to derive the learner's command, where the equivalent teacher command overlaps with the desired teacher command, demonstrating the effectiveness of our transfer framework between the teacher and the learner. Finally, row (d) shows the final frame of the learner that uses the proposed framework.

\begin{figure}[h]
            \centering
            \includegraphics[width=1.0\textwidth]{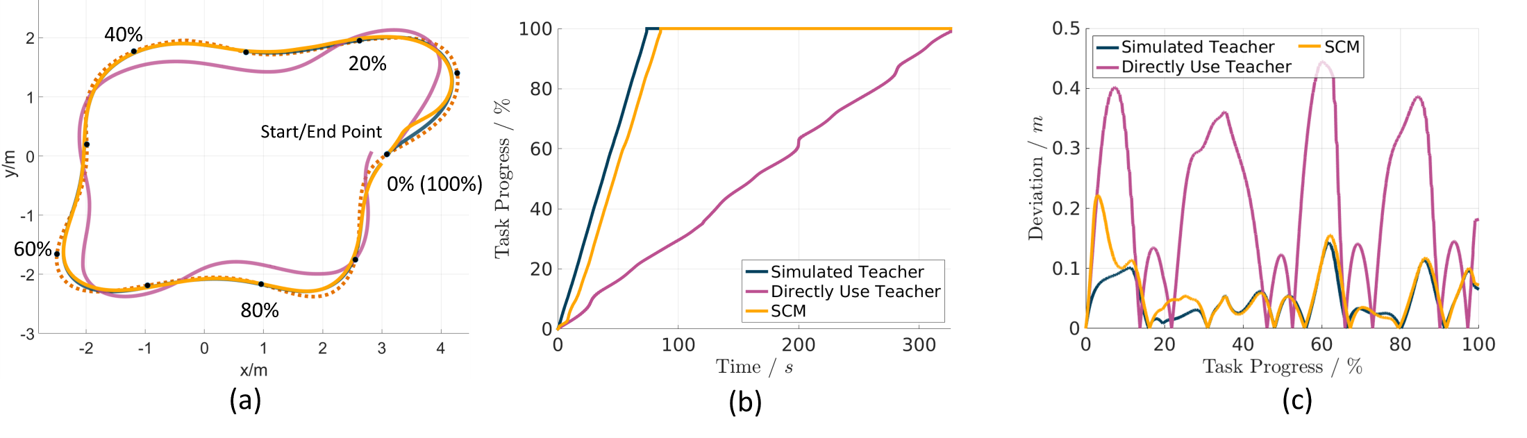}
            \caption{Figure (a) shows a comparison of the progression over time among the proposed approach, the baseline, and the ideal teacher; (b) compares the deviation from the desired path in relation to the task's progression.}
            \label{fig:SCM_MPC_sim_result_analyze}
        \end{figure}

    Figure~\ref{fig:SCM_MPC_sim_result_analyze}(b) shows a comparison of the time taken to complete the same path-tracking task.  Notably, the learner using our proposed approach completes the task shortly after the ideal teacher, while the learner that directly applies the teacher's commands is still in the early stages of the task. This comparison illustrates that our transfer framework enables the learner to not only closely mimic the teacher's maneuvers but also to closely match the teacher's performance. Figure~\ref{fig:SCM_MPC_sim_result_analyze}(c) compares the deviation from the desired path throughout the tracking progress. For ease of comparison, we present Figure~\ref{fig:SCM_MPC_sim_result_analyze}(a) as a reference for task progress indicated as black dots along the tracking path. The errors from the learner with our approach closely align with those of the ideal teacher. Larger deviations occur only when the path requires commands from an unconfigured area of the command domain. Once the learner constructs new command pairs in that area, the deviation significantly decreases, effectively aligning with the teacher's performance. In contrast, the baseline learner shows dramatic differences compared with the ideal teacher.

\begin{figure}
            \centering
            \includegraphics[width=1.0\textwidth]{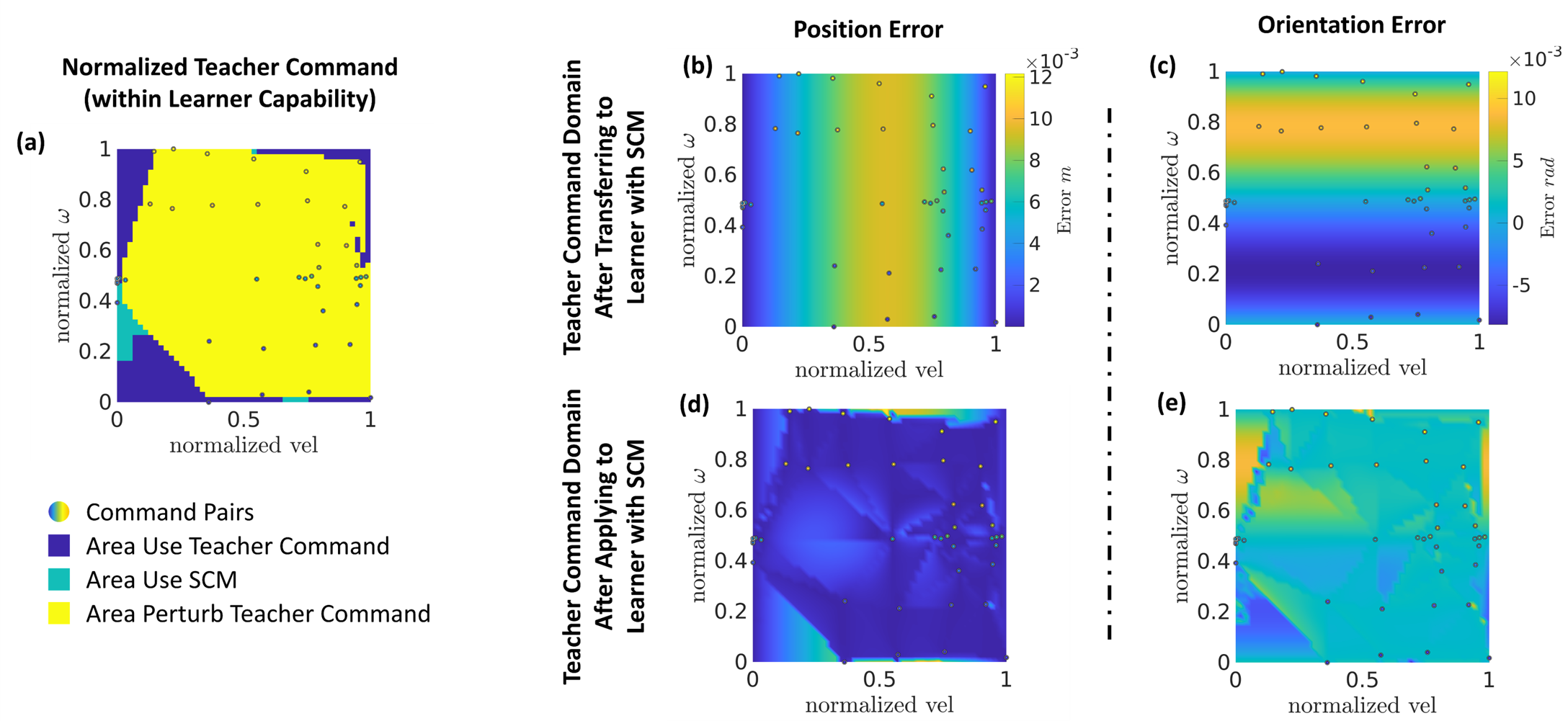}
            \caption{Results from extensive testing of SCM-based command transfer: (a) Normalized teacher command domain (min-max normalization within the learner's limits); (b) Position errors between the teacher and the baseline learner; (c) Orientation errors between the teacher and the baseline learner; (d) Position errors between the teacher and the SCM-enhanced learner; (e) Orientation errors between the teacher and the SCM-enhanced learner, all given the teacher commands and drive the systems for $0.1$s.}
            \label{fig:SCM_MPC_sim_result_error}
        \end{figure}

        To demonstrate the effectiveness of our transfer framework, based on the command pairs and characterized learner limits from the simulation result, we did an exhaustive test of transferring a dense grid of teacher's commands within the learner's limits. The test compared the ideal simulated teacher, the baseline learner, and the learner with transfer method. Given the same teacher command, all three systems start at the same initial pose and drive for $0.1s$. We measure the position errors as well as the orientation errors between the teacher and the two learners. The results are depicted in Figure~\ref{fig:SCM_MPC_sim_result_error}. In the figure, (a) categorizes the normalized teacher command space based on the method that the learner with SCM employed: SCM mapping, direct application of teacher commands, or perturbation for exploring unconfigured spaces. (b) and (d) present the position errors between the ideal teacher and the learners, while (c) and (e) show the orientation errors. The results reveal minimal heading and position errors in areas using SCM mapping (corresponding to the yellow area in (a)), suggesting maneuvers nearly identical to the teacher's. In areas with sparser command pairs, the construction of the mapping polygon utilizes command pairs that are distant from the desired teacher command, resulting in slightly higher errors as expected due to a lack of local geometrical information. Conversely, the baseline learner struggled to match the teacher's behaviors showing larger errors in both position and orientation compared with the ideal teacher.
        

    \FloatBarrier
    \vspace{5pt}
    \noindent
    \textbf{\emph{Experiment Results}}\label{subsec:mpc_scm_exp_results}

\begin{figure}[h]
        \centering
        \begin{minipage}{0.3\textwidth}
            \hfill
            \includegraphics[width=0.8\linewidth]{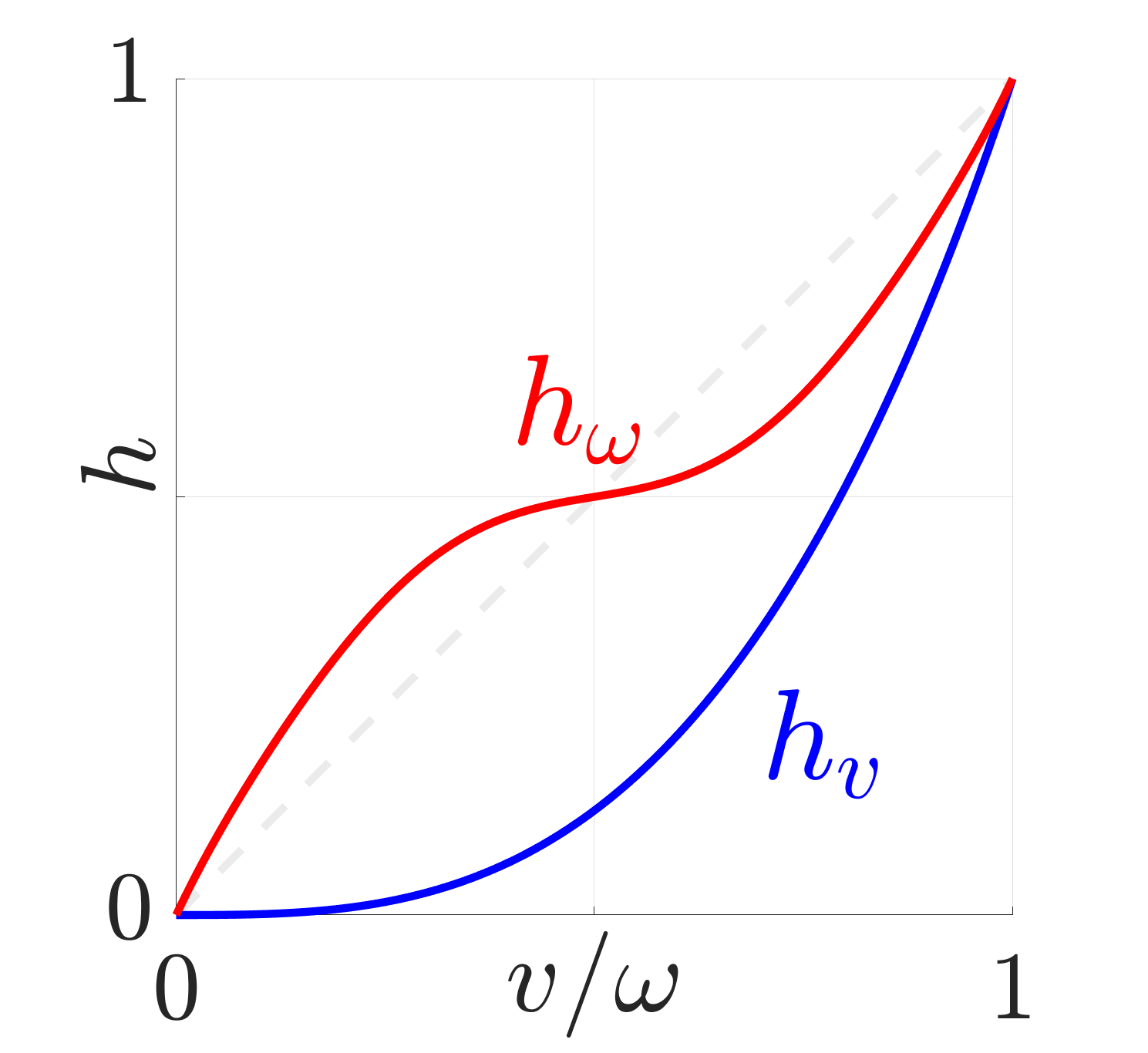}
            \caption{The nonlinear functions employed to alter the learner's dynamics in experiments.}
            \label{fig:nonlinear_exp}
        \end{minipage}%
        \begin{minipage}{0.7\textwidth}
            \begin{equation}
                \begin{aligned}
                    & h_v(v) =  v^3 \\
                    & h_\omega(\omega) = 31.42\omega^7 -109.91\omega^6 - 144.65\omega^5 -86.98\omega^4 \\ &\quad\quad\quad\quad + 24.77\omega^3 -5.08\omega^2 + 2.12\omega \\
                \end{aligned}
                \label{eq:nonlinear_exp}
            \end{equation}
        \end{minipage}
    \end{figure}

    The SCM-based transfer framework with MPC was also validated with real vehicles. The hardware setup is similar to the previous study case in~\ref{subsubsec:primitive_result_exp}. The parameters used for the simulated teacher and the learner vehicle, Clearpath Robotics Jackal, are listed in Table~\ref{tb:scm_mpc_settings}. The learner's command domain is divided into $11\times11$ grid cells with $k_{min}=20$. The nonlinear transformation functions used to further alter the Jackal's kinematics are described in Eq.~\eqref{eq:nonlinear_exp} and their visualizations are depicted in Figure~\ref{fig:nonlinear_exp}.
            

    
\begin{figure}[h]
            \centering
            \includegraphics[width=0.85\linewidth]{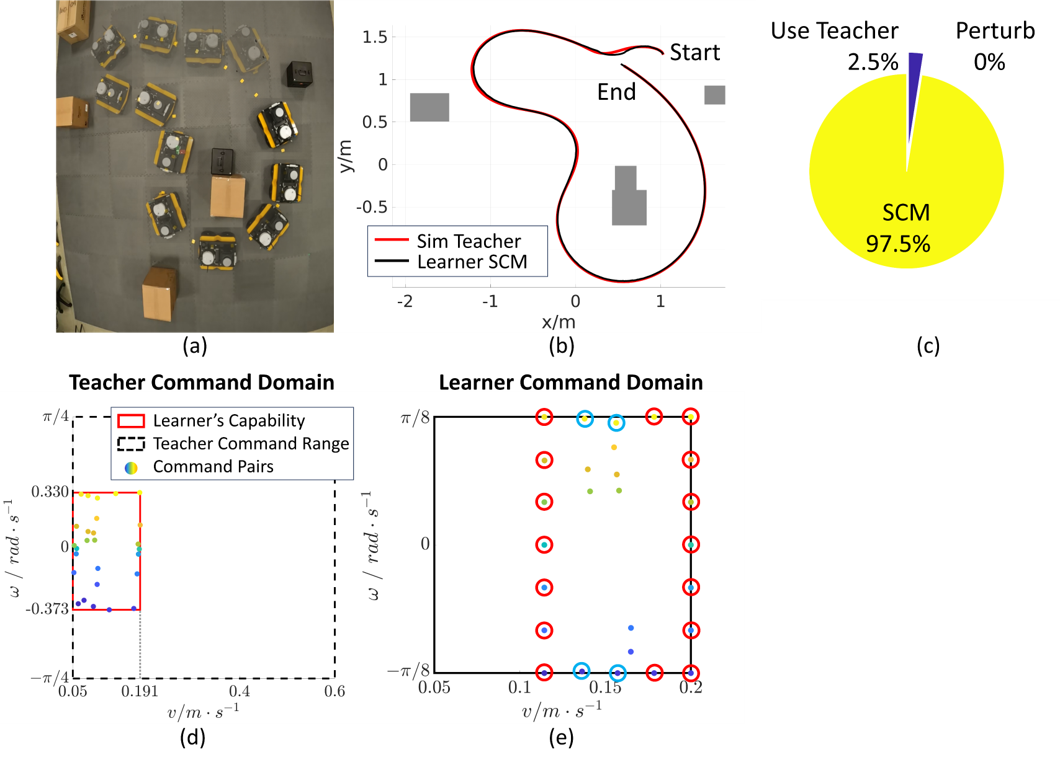}
            \caption{(a) snapshots of the experiment; (b) compares the trajectory between the ideal teacher in simulation and the learner with the proposed approach; \REV{(c) summarizes the method used for obtaining the final learner's command; (d) and (e) present the color-coded command pairs on the teacher's and the learner's command domain respectively after following the desired path.}}
            \label{fig:SCM_MPC_exp_result}
        \end{figure}

\begin{figure}[h]
            \centering
            \includegraphics[width=\linewidth]{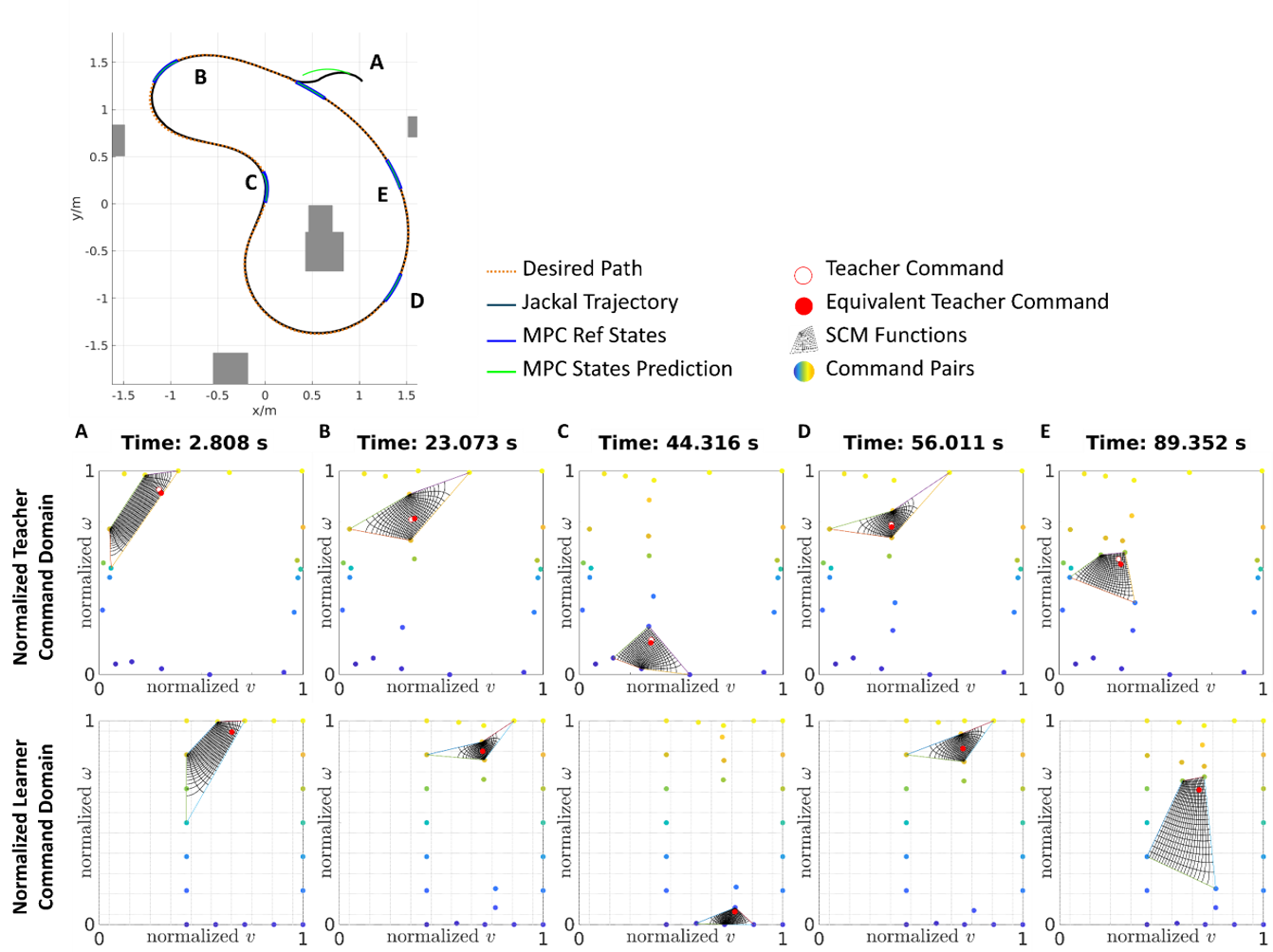}
            \caption{Examples from experiment demonstrating the use of SCM to transfer MPC from a simulated teacher to a Clearpath Jackal UGV at different time frames.}
            \label{fig:SCM_MPC_exp_process}
        \end{figure}

    Due to limited indoor space, a set of command pairs characterizing the Jackal's limits was pre-constructed. Figure~\ref{fig:SCM_MPC_exp_result} shows the results of a path-tracking experiment with the Jackal, with snapshots provided in Figure~\ref{fig:SCM_MPC_exp_process}. The Jackal starts positioned away from the desired path. \REV{In Figure~\ref{fig:SCM_MPC_exp_result}(e), the pre-constructed command pairs are circled, with those updated during the experiment highlighted in blue.} Most commands sent to the learner are derived through conformal mapping, while a few low-velocity commands are directly applied or perturbed from the teacher's commands due to the absence of a suitable polygon for SCM application.
    

\begin{figure}[h]
            \centering
            \subfigure[]{\includegraphics[height=4cm,keepaspectratio]{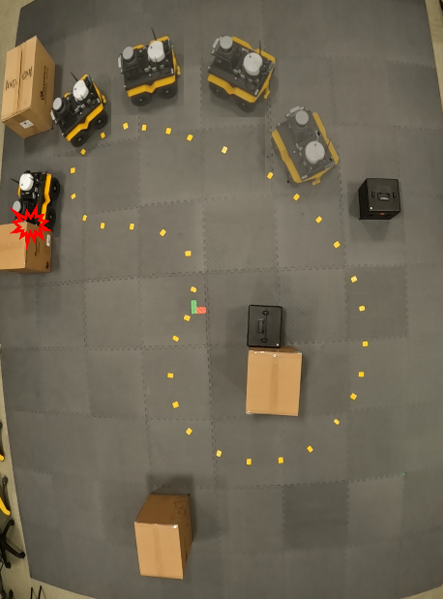}\label{fig:noscm_mpc_exp}}
            \subfigure[]{\includegraphics[height=3.5cm,keepaspectratio]{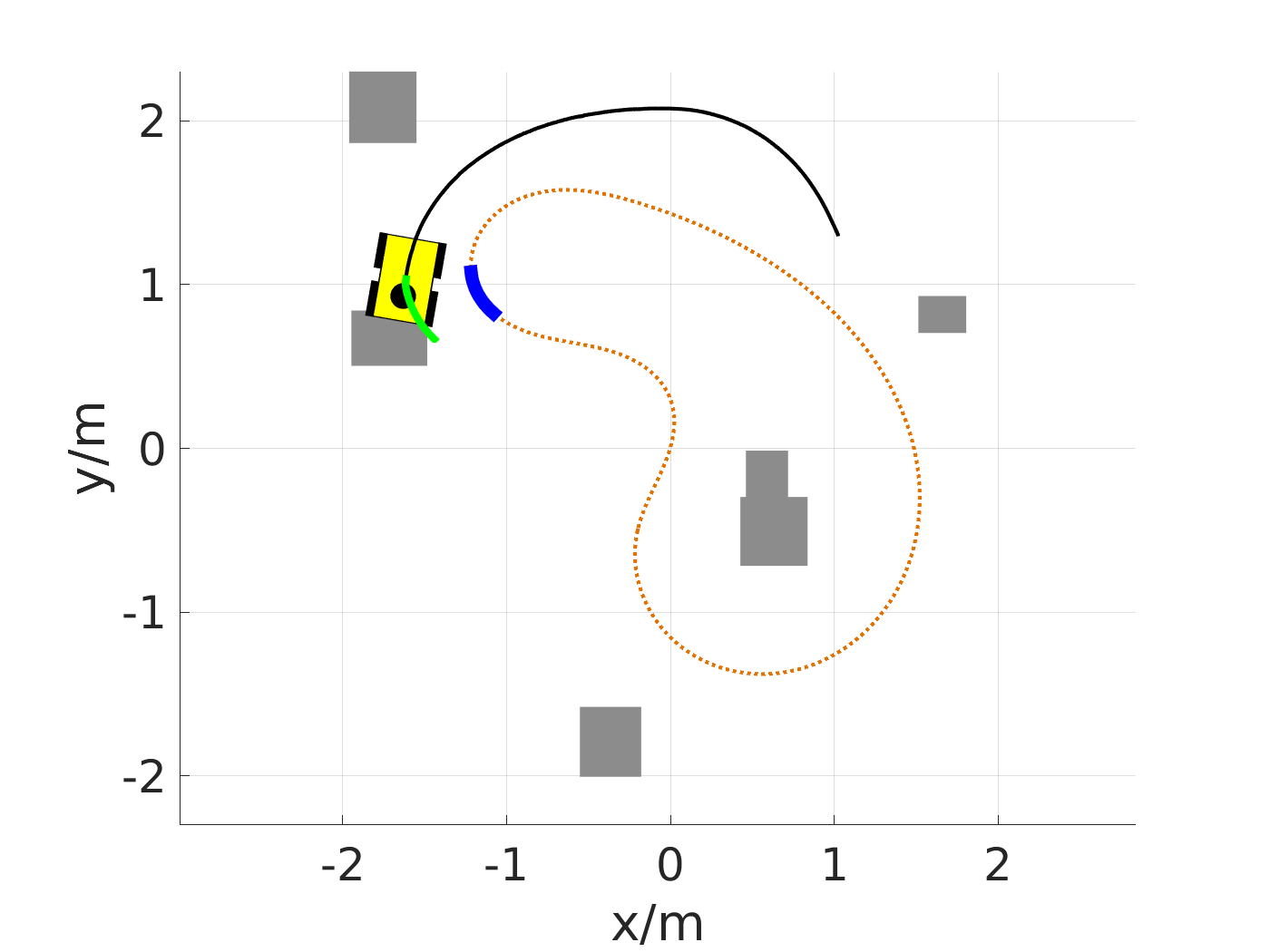}\label{fig:noscm_mpc_exp_traj}}
            \caption{The result of directly applying the simulated teacher's commands to the Jackal}
            \label{fig:noSCM_MPC_exp}
        \end{figure}

\begin{figure}[h]
            \centering
            \includegraphics[width=.9\linewidth]{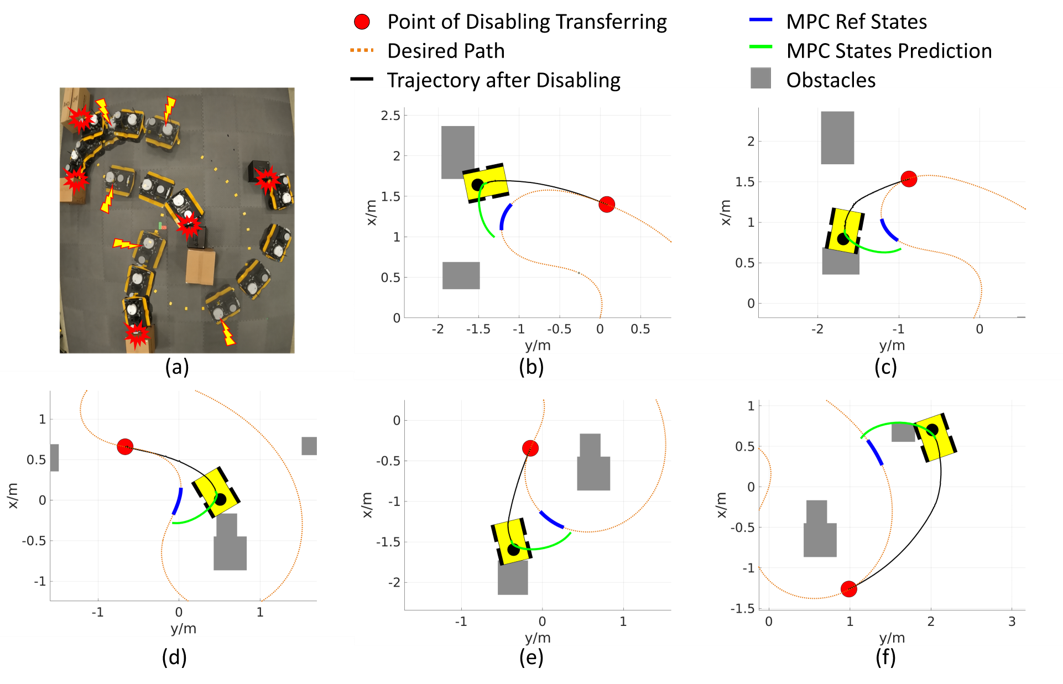}
            \caption{Figure (a) presents the result of deactivating the proposed approach and directly applying the teacher's command at different points during the task; (b)-(f) show the learner's trajectory, the reference states, and the predicted states from the teacher's MPC controller at the moment the learner crashes into obstacles.}
            \label{fig:SCM_MPC_toggle_exp}
        \end{figure}

    The effectiveness of the proposed transferring framework is further validated by comparison with a baseline Jackal for the same task, where teacher commands are directly applied, as shown in Figure~\ref{fig:noSCM_MPC_exp}. The baseline Jackal moved much slower, failed to follow the path and ultimately collided with an obstacle. In a separate test, we deactivated the transfer framework at various points during the task, which led to the direct application of the teacher's control inputs after the deactivation. Figure~\ref{fig:SCM_MPC_toggle_exp} shows the Jackal's trajectory after the framework is turned off, and in all five cases, the Jackal deviated from the desired path colliding with obstacles.

\section{Conclusion and Discussion}\label{sec:conclusion}
In this work, we have introduced a novel, lightweight transfer learning framework based on conformal mapping, designed to bridge the sim-to-real gap and address model mismatches. Our transfer framework achieves control transfer in both discrete and continuous command space, exemplified by successfully transferring two representative methods (MPC and motion primitive-based motion plan) in autonomous mobile robot applications. We enhanced the robustness and generalizability of our proposed method by 1) clustering the command pairs to address the challenge of constructing command pairs in the presence of unneglectable motion noise; 2) actively refining the learner’s limits along with the transferring process, eliminating the need for a dedicated calibration period. The effectiveness of the framework is validated with extensive simulations and real-world experiments. Videos of the simulations and experiments are provided in the supplementary material and can be accessed at {\url{https://www.bezzorobotics.com/sg-jint24}}.

Our framework achieves seamless behavior transfer between a known teacher system and an unknown learner system by directly transferring control inputs without needing to precisely model the learner. The control transfer within our framework is based on command pairs that elicit identical maneuvers in both the teacher and learner systems. It operates by first learning the geometrical distribution of these command pairs, and then employing the Schwarz-Christoffel Mapping method to map between these pairs. This mapping accurately locates the corresponding learner command for a given teacher command, ensuring the desired maneuver is replicated in the learner vehicle. Conformal mapping is well-established in two-dimensional spaces but becomes considerably more complex and limited in higher dimensions. Therefore, we concentrate on systems whose input spaces are inherently two-dimensional or can be bijectively represented in two dimensions. Since SCM is supported by rich and robust mathematical theories, our design choices are driven by mathematical principles to optimize the performance of the mapping function calculation. We pick out a few key design points that are worth mentioning here:
    \begin{enumerate}
        \item \textbf{Using bi-infinite strip in rectangle SCM.}
        ``Crowding" is one of the greatest challenges for numerically computing conformal mapping functions. The high ratio of an elongated shape can lead to a situation where the prevertices are spaced exponentially close on the real axis becoming indistinguishable. A bi-infinite strip can significantly ease the effort of solving numerical SCM solutions over the elongated shape. Even for a less elongated shape, this can speed up the computing process. As mathematically solving the SCM is not the main contribution of this work, we direct readers to refer~\cite{driscoll2002schwarz,howell1990schwarz} for a qualitative analysis of how using bi-infinite strip eases rectangle SCM computation. 
        
        \item \textbf{The reason of choosing rectangle SCM.}
        While our proposed transfer framework primarily utilizes rectangle SCM, which maps polygon regions with at least four vertices ($N \geq 4$), it is also compatible with triangle SCM ($N =3$). With triangle SCM, triangles from both command domains can be mapped directly to the same set of prevertices on a disk or an upper half-plane, simplifying the process by eliminating the need to connect both ends with a unit disk. However, rectangle SCM is preferred in practice because it can incorporate more command pairs without as much concern for SCM crowding issues.

        \item \textbf{The distribution density of the command pairs.}
        Our proposed approach directly transfers the control input by geometrically mapping across the command domains. The nature of this approach depends on accurate command pairs and benefits from selecting pairs that are geometrically close to the desired command. Closer command pairs are more effective as they capture local geometric information more accurately and better reflect the learner system's similar motions.
        Thus, having the mapping area well covered by the command pairs is advantageous.

        \item\textbf{\REV{Transferring between high-dimensional($\geq3$) control spaces.}}
        \REV{The approach presented in this work aims to demonstrate the concept of leveraging conformal mapping to directly transfer control inputs between systems with two-dimensional control spaces. By following the same principle, conformal mapping methods in high-dimensional spaces could potentially be used to directly transfer control inputs between systems with high-dimensional control spaces. One alternative for systems with high-dimensional control spaces is Liouville’s Theorem (LT)\cite{phillips1969liouville}, a well-established theory for conformal mappings in spaces of dimension three or higher. However, LT also states that in Euclidean spaces of dimension greater than two, conformal mappings must, by necessity, be Möbius transformations (i.e., a translation, a magnification, an orthogonal transformation, a reflection through reciprocal radii, or a combination of these transformations). This restriction indicates that the potential for using multidimensional conformal mapping to transfer control inputs may be more limited than in the two-dimensional case, given the wealth of conformal maps available in the plane. Therefore, if possible, we encourage leveraging a bijective representation of high-dimensional systems in two-dimensional spaces to fully benefit from the richness of 2D conformal mapping.}
        
    \end{enumerate}


\REV{Moving forward, having demonstrated that conformally transferring control inputs between different control domains can enable smooth sim-to-real transitions, we plan to compare our proposed approach with other state-of-the-art sim-to-real transfer methods.} We also aim to expand our proposed framework to facilitate command transfer between heterogeneous robotic systems, addressing and mitigating more challenging model discrepancies. Potential future directions include expanding our proposed framework to facilitate knowledge transfer between a limited teacher and more capable learners. In scenarios where learners possess greater capabilities, our framework can effectively align the learner's performance with that of a limited teacher, serving as a baseline, while employing additional techniques to harness and maximize the learner's capabilities. Additionally, we are interested in utilizing SCM to transfer control and planning policies from low-dimension to high-fidelity representations.

\section{Acknowledgments}
This material is based upon work supported by the Defense Advanced Research Projects Agency (DARPA) under Grant No. FA8750-18-C-0090. Any opinions, findings, conclusions, or recommendations expressed in this material are those of the authors and do not necessarily reflect the views of the DARPA. 

\section*{Declarations}
\begin{itemize}
    \item \textbf{Competing Interests:} The authors have no financial interest to declare.
    \item \textbf{Author Contributions:} All authors contributed to the study conception and design. The research investigation, methodology development, material preparation, and initial manuscript drafting were conducted by Shijie Gao. Nicola Bezzo provided supervision for the study and contributed to the manuscript review and editing. All authors have read and approved the final version of the manuscript.
    \item \textbf{Ethics Approval:} Not applicable.
    \item \textbf{Consent to Participate:} Not applicable.
    \item \textbf{Consent to Publish:} Not applicable.
\end{itemize}


\bibliography{sn-bibliography}

\end{document}